\documentclass[letterpaper]{article} 
\usepackage{aaai24}  
\usepackage{times}  
\usepackage{helvet}  
\usepackage{courier}  
\usepackage[hyphens]{url}  
\usepackage{graphicx} 
\usepackage{natbib}  
\usepackage{caption} 
\usepackage{algorithm}
\usepackage{algorithmic}

\usepackage[dvipsnames]{xcolor}
\usepackage{natbib}
\usepackage{graphicx}
\usepackage{amsmath,amssymb}
\usepackage{multirow}
\usepackage{color}
\usepackage{amssymb,mathrsfs}
\usepackage{subfigure}
\usepackage{amsthm}
\usepackage{booktabs}

\usepackage{newfloat}
\usepackage{listings}
\DeclareCaptionStyle{ruled}{labelfont=normalfont,labelsep=colon,strut=off} 
\floatstyle{ruled}
\newfloat{listing}{tb}{lst}{}
\floatname{listing}{Listing}
\newtheorem{theorem}{Theorem}
\newtheorem{corollary}{Corollary}
\newtheorem{assumption}{Assumption}

\newtheorem{proposition}{Proposition}
\newtheorem{premise}{Premise}

\title{AdapterGNN: Parameter-Efficient Fine-Tuning Improves Generalization in GNNs}
\author{
    Shengrui Li\textsuperscript{\rm 1,\rm 2}\thanks{Work is done during internship at Microsoft Research Asia.}, 
    Xueting Han\textsuperscript{\rm 2}\thanks{Corresponding author.},
    Jing Bai\textsuperscript{\rm 2},
}
\affiliations{
    \textsuperscript{\rm 1}Tsinghua University\\
    
    \textsuperscript{\rm 2}Microsoft Research Asia\\

    lsr22@mails.tsinghua.edu.cn, \{chrihan, jbai\}@microsoft.com\\

}

\usepackage{bibentry}

\begin{document}

\newcommand{\tmprefdelta}[1]{C.1}
\newcommand{\tmpreffinite}[1]{B.1}
\newcommand{\tmprefsup}[1]{B.2}
\newcommand{\tmprefpre}[1]{B.3}
\newcommand{\tmprefpreserve}[1]{C.2}
\newcommand{\tmprefimplementation}[1]{D.2}
\newcommand{\tmprefdataset}[1]{D.3}
\newcommand{\tmprefpretrain}[1]{D.4}
\newcommand{\tmprefdetail}[1]{D.1}

\newcommand{\tmprefemb}[1]{6}
\newcommand{\tmprefcolsup}[1]{1}
\newcommand{\tmprefcolpre}[1]{2}
\newcommand{\tmprefinequa}[1]{5}
\newcommand{\tmpreffigabl}[1]{4}
\newcommand{\tmprefflops}[1]{C.3}

\maketitle

\begin{abstract}
Fine-tuning pre-trained models has recently yielded remarkable performance gains in graph neural networks (GNNs). In addition to pre-training techniques, inspired by the latest work in the natural language fields, more recent work has shifted towards applying effective fine-tuning approaches, such as parameter-efficient fine-tuning (PEFT). However, given the substantial differences between GNNs and transformer-based models, applying such approaches directly to GNNs proved to be less effective. In this paper, we present a comprehensive comparison of PEFT techniques for GNNs and propose a novel PEFT method specifically designed for GNNs, called AdapterGNN. AdapterGNN preserves the knowledge of the large pre-trained model and leverages highly expressive adapters for GNNs, which can adapt to downstream tasks effectively with only a few parameters, while also improving the model's generalization ability. Extensive experiments show that AdapterGNN achieves higher performance than other PEFT methods and is the only one consistently surpassing full fine-tuning (outperforming it by 1.6\% and 5.7\% in the chemistry and biology domains respectively, with only 5\% and 4\% of its parameters tuned) with lower generalization gaps. Moreover, we empirically show that a larger GNN model can have a worse generalization ability, which differs from the trend observed in large transformer-based models. Building upon this, we provide a theoretical justification for PEFT can improve generalization of GNNs by applying generalization bounds. Our code is available at https://github.com/Lucius-lsr/AdapterGNN.
\end{abstract}

\section{Introduction}

Graph neural networks (GNNs) \cite{scarselli2008graph, wu2020comprehensive} have achieved remarkable success in analyzing graph-structured data  \cite{hamilton2017inductive, velickovic2017graph, xu2018powerful} but face challenges such as the scarcity of labeled data and low out-of-distribution generalization ability. To overcome these challenges, recent efforts have focused on designing GNNs pre-training approaches \cite{hu2019strategies, xia2022simgrace, you2020graph} that leverage abundant unlabeled data to capture transferable intrinsic graph properties and generalize them to different downstream tasks by fine-tuning \cite{zhou2019meta,zhou2019auto,wang2022advanced}. While fine-tuning all parameters from a pre-trained model can improve performance \cite{peng2019using, hu2019strategies}, it usually requires a relatively large model architecture to effectively glean knowledge from pre-training tasks \cite{devlin2018bert, brown2020language}. This becomes challenging when the downstream task has limited data, as optimizing a large number of parameters can lead to overfitting \cite{lecun2015deep}. Moreover, training and maintaining a separate large-scale model for each task can prove to be inefficient as the number of tasks grows. To address these challenges, recent research has focused on developing parameter-efficient fine-tuning (PEFT) techniques that can effectively adapt pre-trained models to new tasks \cite{ding2022delta}, such as adapter tuning \cite{houlsby2019parameter}, LoRA \cite{hu2021lora}, BitFit \cite{zaken2021bitfit}, prefix-tuning \cite{li2021prefix}, and the prompt tuning \cite{lester2021power}. PEFT seeks to tune a small portion of parameters and keep the left parameters frozen. This approach reduces training costs and allows for use in low-data scenarios.
PEFT could be applied to the GNNs \cite{xia2022survey}. In particular, the idea of prompt tuning has been widely adopted to GNNs \cite{wu2023survey}. This adoption involves either manual prompt engineering \cite{sun2022gppt} or soft prompt tuning techniques \cite{liu2023graphprompt, diao2022molcpt, fang2022prompt}. However, prompt-based methods involve modifying only the raw input and not the inner architecture and thus struggling to match full fine-tuning performance \cite{fang2022prompt}, unless in special few-shot setting \cite{liu2023graphprompt}. Existing works lack study and comparison with other PEFT methods. And due to the inherent difference between transformer-based models and GNNs, not all NLP solutions can be directly applied to GNNs.

\begin{figure}[t]
\centering
\includegraphics[width=0.9\columnwidth]{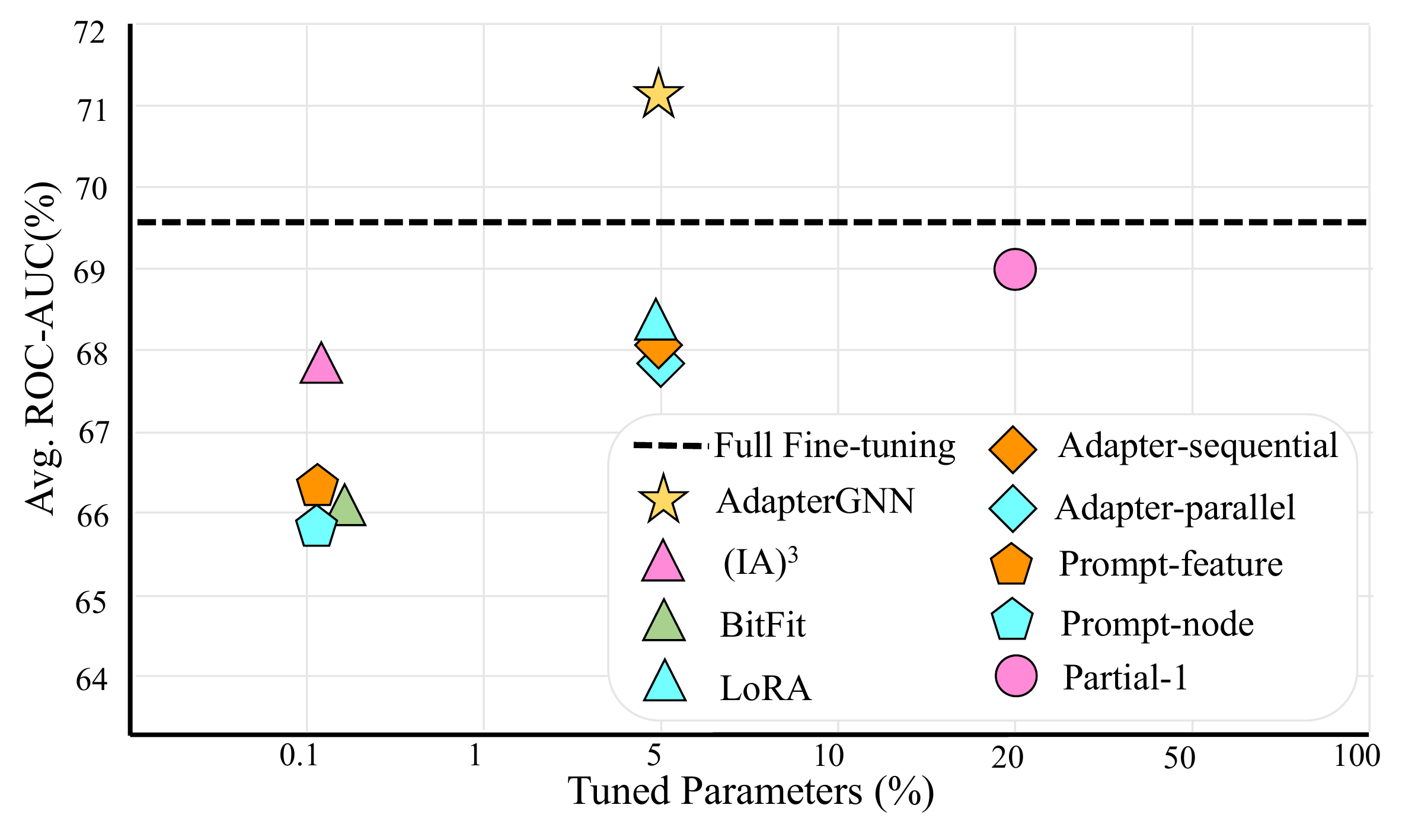}
    \caption{Comparison among various PEFT methods of GNNs in six small molecular datasets. Detailed comparisons are shown in Table \ref{table_main} and explanations are in Appendix \tmprefimplementation.}
  \label{fig_compare}
\end{figure}

To address this issue, we propose an effective method, AdapterGNN, that combines task-specific knowledge in tunable adapters with task-agnostic knowledge in the frozen pre-trained model. Unlike the vanilla adapter \cite{houlsby2019parameter} which exhibits poor performance when directly applied to GNNs, AdapterGNN is specifically designed to cater to the non-transformer GNN architecture employing novel techniques: (1) Dual adapter modules; (2) Batch normalization (BN); (3) Learnable scaling.

As illustrated in Fig. \ref{fig_compare}, we conduct a comprehensive comparison of various PEFT approaches in GNNs. Applying them on GNNs is non-trivial since most of them were implemented on transformer-based models. AdapterGNN, with only 5\% tuned parameters, achieves higher evaluation performance than other PEFT methods and is the only one consistently surpassing full fine-tuning.

This improvement can be attributed to AdapterGNN maximally utilizing advantages of PEFT with special designs. PEFT addresses two drawbacks of full fine-tuning to improve generalization. First, catastrophic forgetting of pre-trained knowledge is disastrous during generalizing \cite{kirkpatrick2017overcoming}. Since PEFT keeps most parameters fixed, catastrophic forgetting can be potentially mitigated, resulting in improved model transferability and generalization \cite{ding2022delta}. 
Second, overfitting is severe when tuning on a small dataset with large parameters \cite{aghajanyan2020intrinsic, arora2018stronger}, particularly in the OOD case \cite{kuhn2013over, hu2019strategies}. 

For the first time, we provide a detailed theoretical justification from the perspective of generalization bounds \cite{mohri2018foundations, shalev2014understanding} to explain how PEFT in GNNs mitigates overfitting and promotes generalization ability. 

To conclude, our work makes the following contributions:
\begin{itemize}
\item We are the first to apply the adapter to GNNs. To cater to the non-transformer GNN architecture, we integrate special techniques (dual adapter modules, BN, learnable scaling), which are essential in improving performance.
\item We are the first to provide a theoretical justification of how PEFT improves generalization for non-transformer models. While there are works to study the theoretical support of PEFT in large transformer models, there is a lack of theoretical study for non-transformer models, e.g., GNNs. Although numerous empirical studies have demonstrated the efficacy of PEFT, we contribute to filling the gap between theory and empirical results.
\item We are the first to apply all prevalent PEFT methods on GNNs and provide a detailed comparison. Applying such PEFT methods on GNNs is non-trivial since most PEFT methods were implemented on transformer-based models. We make special variations and implement them in GNNs, which fills the void in this field.
\end{itemize}

\section{Related Work}

\paragraph{Parameter-efficient fine-tuning methods.}
Full fine-tuning tunes all the model parameters and adapts them to downstream tasks, but this becomes inefficient with the growth of model size and task count. Recent NLP work has explored PEFT techniques that tune only a small portion of parameters for efficiency \cite{ding2022delta}. Prompt tuning \cite{lester2021power} aims to modify model inputs rather than model architecture. Prefix-tuning \cite{li2021prefix} only updates task-specific trainable parameters in each layer. Adapter tuning \cite{houlsby2019parameter, chen2022adaptformer} inserts adapter modules with bottleneck architecture between layers. BitFit \cite{zaken2021bitfit} only updates the bias terms while freezing the remaining. LoRA \cite{hu2021lora} decomposes the weight matrix into low-rank matrices to reduce the number of trainable parameters.
As for the GNNs field, the idea of prompt tuning has gained widespread acceptance \cite{wu2023survey}. GPPT \cite{sun2022gppt} specially designs a framework for GNNs but is limited to node-level tasks. Limited to the molecular field, MolCPT \cite{diao2022molcpt} encodes additional molecular motif information to enhance graph embedding. GPF \cite{fang2022prompt} and GraphPrompt \cite{liu2023graphprompt} are parameter-efficient but struggle to match the full fine-tuning baseline in the non-few-shot setting.

\paragraph{Generalization error bounds.}

Generalization error bounds, also known as generalization bounds, provide insights into the predictive performance of learning algorithms in statistical machine learning. In the classical regime, bias–variance tradeoff states that the test error as a function of model complexity follows the U-shaped behavior \cite{mohri2018foundations, shalev2014understanding}. However, in the over-parameterized regime, increasing complexity decreases test error, following the modern intuition of larger model generalizes better \cite{belkin2019reconciling, nakkiran2021deep, zhang2021understanding, sun2016depth, hardt2016train, mou2018generalization}. \cite{aghajanyan2020intrinsic} has theoretically analyzed generalization on over-parameterized large language models, it explains the empirical results that larger pre-trained models generalize better by applying intrinsic-dimension-based generalization bounds. While our work utilizes conventional generalization bounds to analyze the generalization ability of PEFT on GNNs. 

\section{Parameter-Efficient Fine-Tuning Improving Generalization Ability in GNNs}\label{sec_prove}

\begin{figure}[t]
\centering
\includegraphics[width=0.9\columnwidth]{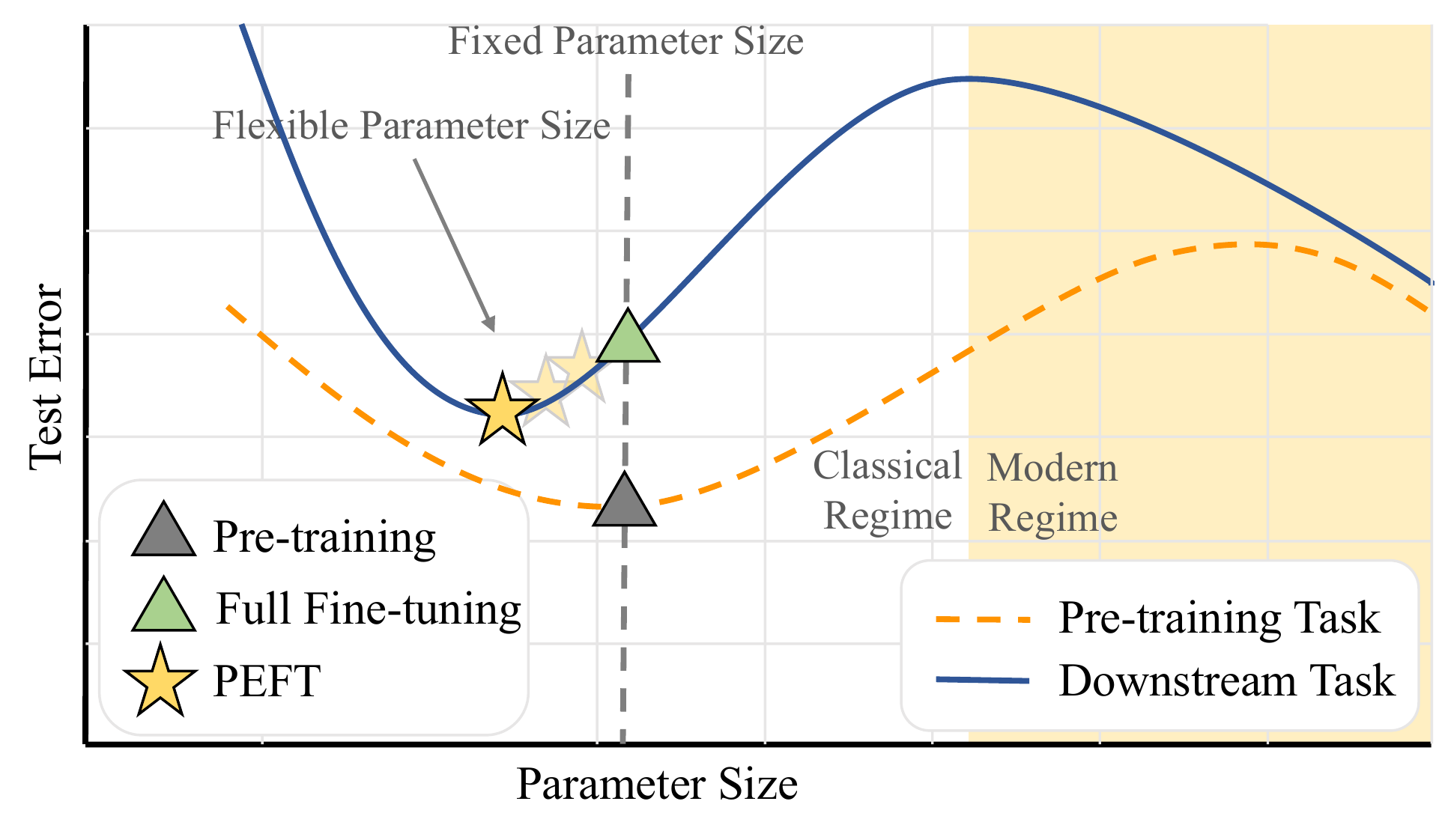}
    \caption{A large model is often employed for pre-training \textcolor{Gray}{$\blacktriangle$} when sufficient data is available. However, for downstream tasks with limited data, a smaller model is optimal in the classical regime. Compared with full fine-tuning \textcolor{LimeGreen}{$\blacktriangle$}, PEFT \textcolor{Goldenrod}{$\bigstar$} preserves expressivity while reducing the size of parameter space, leading to lower test error.}
  \label{fig_principle}
\end{figure}


Several works have shown that transformer-based large models in NLP/CV are over-parameterized and empirically shown larger pre-trained models generalize better,  aligning with the modern regime in Fig. \ref{fig_principle}. They use generalization bounds which are independent of the parameter count for large pre-trained models \cite{aghajanyan2020intrinsic}.

However, theoretical study for PEFT in models aligning with the classical regime is still lacking. Our empirical findings in Appendix \tmprefdelta{app_delta} find that larger GNNs generalize worse, satisfying the classical regime in Fig. \ref{fig_principle}. Therefore, we explore how PEFT benefits these models. We theoretically demonstrate PEFT can lower the bounds of test error (generalization bounds) and improve generalization ability in GNNs compared to full fine-tuning, as illustrated in Fig. \ref{fig_principle}. Our justification begins with this premise:

\begin{premise}
GNNs satisfy the classical regime of generalization bounds theory.
\end{premise}

Within this regime, we apply a widely used parameter counting based generalization bounds theorem \cite{aghajanyan2020intrinsic, arora2018stronger} and the detailed proof of this theorem can be found in Appendix \tmpreffinite{app_finite}:





\begin{theorem} \label{theo1} \textbf{Generalization bounds for finite hypothesis space in classical regime.} Training data $\mathcal{D}_n$ and the trained parameters $\mathcal{P}$ are variables of training error $\hat{\mathcal{E}}$. The number of training samples $n$ and size of parameter space $|\mathcal{P}|$ are variables of generalization gap bounds. Then, statistically, the upper bound $\mathcal{U}(\mathcal{E})$ of the test error $\mathcal{E}$ of a model in finite hypothesis space is determined as follows:
\begin{equation}
    \mathcal{E} \leq\mathcal{U}(\mathcal{E})=\hat{\mathcal{E}}({\mathcal{D}_n},\mathcal{P})+O\left(\sqrt{|\mathcal{P}|/n}\right).
\end{equation}
\end{theorem}
 Before introducing PEFT, we first compare the error bounds of two paradigms: "supervised training from scratch" and "pre-train, fine-tune". For supervised training from scratch, we denote the task as $T$ and training data as $\mathcal{D}_{n_T}^T$. With the increase of $|\mathcal{P}_T|$, training error $\hat{\mathcal{E}}(\mathcal{D}_{n_T}^T, \mathcal{P}_T)$ decreases due to stronger optimization capability and generalization gap bounds $O(\sqrt{|\mathcal{P}_T|/n_T})$ increases. Therefore, following the U-shaped behavior, the following corollary is obtained and the detailed proof can be found in Appendix \tmprefsup{app_sup}:

\begin{corollary} \textbf{Bounds of supervised training from scratch.}
There is an optimal $| \bar{\mathcal{P}_T|}$ to 
get the tightest upper bound $min(\mathcal{U}(\mathcal{E}_T))=\hat{\mathcal{E}}({\mathcal{D}_{n_T}^T},\bar {\mathcal{P}_T})+O\left(\sqrt{|\bar {\mathcal{P}_T}|/n_T}\right)$.
\label{col_sup}
\end{corollary}
 


For pre-training task $S$, data is more abundant: $n_S>n_T$. Many previous works discover that with abundant data to pre-train, we should employ a larger model to capture sufficient knowledge \cite{han2021pre, ding2022delta, wei2022emergent}. Therefore, we have the following corollary and the proof can be found in Appendix \tmprefpre{app_pre}:


\begin{corollary} \textbf{Bounds of pre-training.}
For the pre-training task $S$ satisfying $n_S>n_T$, there is also an optimal $|\bar {\mathcal{P}_S}|$ and $|\bar{\mathcal{P}_S}|>|\bar{\mathcal{P}_T}|$ to get the tightest upper bound $min(\mathcal{U}(\mathcal{E}_S)) = \hat{\mathcal{E}}({\mathcal{D}_{n_S}^S},\bar {\mathcal{P}_S})+O(\sqrt{|\bar {\mathcal{P}_S}|/n_S})$.
\label{col_pre}
\end{corollary}

In the "pre-train, fine-tune" paradigm, a model that has been pre-trained on task $S$ is used as initialization to improve the performance of a supervised task. To account for the effect of the initial parameter values, we use $\hat{\mathcal{E}_S}(D,\mathcal{P})$ to denote the training error of the pre-trained model.



Compared to training from scratch with the same model as pre-training: $\mathcal{U}(\mathcal{E}_T)= \hat{\mathcal{E}}({\mathcal{D}_{n_T}^T},\bar {\mathcal{P}_S})+O(\sqrt{|\bar {\mathcal{P}_S}|/n_T})$, fine-tuning benefits from a better initialization, which leads to a lower error bound: $\mathcal{U}(\mathcal{E}_F)= \hat{\mathcal{E}}_S({\mathcal{D}_{n_T}^T},\bar {\mathcal{P}_S})+O(\sqrt{|\bar {\mathcal{P}_S}|/n_T})$. Based on this, we make the following assumption and corollary:

\begin{assumption}
The \textbf{T}ransfer \textbf{G}ain $\mathbf{TG}$ from the pre-trained model can be quantified. It is solely determined by the properties of $S, T$ and can be calculated as: $\mathbf{TG} = \hat{\mathcal{E}}({\mathcal{D}_{n_T}^T},\bar P_S) -\hat{\mathcal{E}}_S({\mathcal{D}_{n_T}^T},\bar P_S)$
\end{assumption}

\begin{corollary} \label{col_fine} \textbf{Bounds of full fine-tuning.}
Full fine-tuning of a pre-trained model can result in a lower error bound, which can be measured by the transfer gain denoted by $\mathbf{TG}$:
\begin{equation}\label{equa_fine}
\mathcal{U}(\mathcal{E}_F) = \hat{\mathcal{E}}({\mathcal{D}_{n_T}^T},\bar {\mathcal{P}_S})+O\left(\sqrt{|\bar {\mathcal{P}_S}|/n_T}\right) - \mathbf{TG}
\end{equation}
\end{corollary}

In regard to PEFT, the PEFT model is initialized with the same pre-trained parameters as fine-tuning, so $\mathbf{TG}$ is inherited similarly to fine-tuning, leading to the following corollary:

\begin{corollary} \textbf{Bounds of PEFT.}
 Bounds of PEFT $\mathcal{U}(\mathcal{E}_E)$ share a similar form as full fine-tuning:
\begin{equation}\label{equa_delta}
    \mathcal{U}(\mathcal{E}_E)= \hat{\mathcal{E}}({\mathcal{D}_{n_T}^T}, {\mathcal{P}_E})+O\left(\sqrt{| {\mathcal{P}_E}|/n_T}\right)-\mathbf{TG}.
\end{equation}
\end{corollary}

For PEFT, we propose a prerequisite: the preservation
of the expressivity of full fine-tuning GNNs. Our AdapterGNN is specifically designed to maintain the capacity of the original GNNs maximally. With comparable parameters, it can be almost as powerful as full fine-tuning. We validate this in Appendix \tmprefpreserve{app_preserve}.
In contrast, several PEFT techniques have failed to yield satisfactory results in GNNs, such as prompt tuning \cite{fang2022prompt, liu2023graphprompt}. This can be attributed to not meeting this prerequisite. 

\begin{figure*}[h!]
  \centering
  \includegraphics[width=0.9\textwidth]{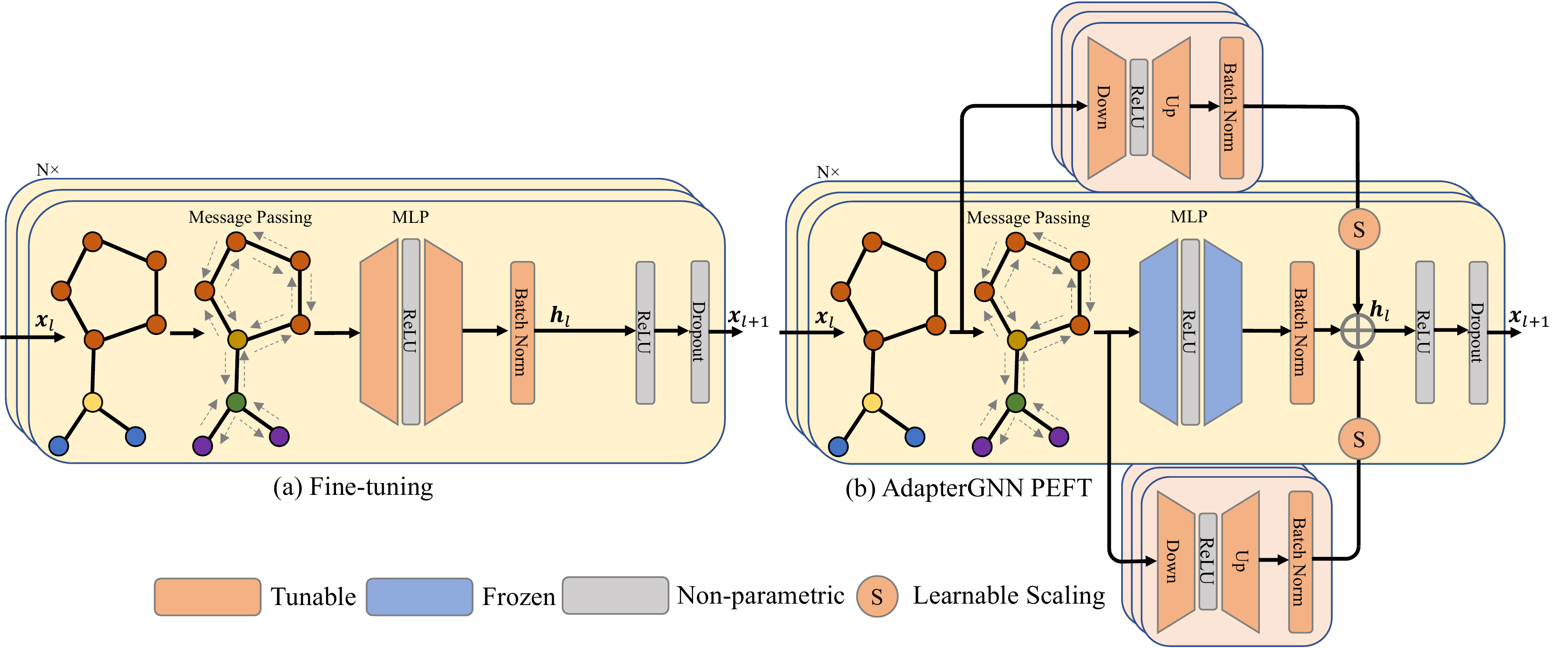}
  \caption{\textbf{Comparison between full fine-tuning and AdapterGNN PEFT on GIN.} a) The full fine-tuning updates all parameters of each pre-trained GNN layer. (b) AdapterGNN includes two parallel adapters taking input before and after the message passing. Their outputs are added to the original output of batch normalization with learnable scaling. During tuning, the original MLP of each GNN layer, which comprises the majority of the parameters, is frozen.}
  \label{fig_framework}
\end{figure*}

With this prerequisite, when training from scratch on $T$, the trainable structure of PEFT and fine-tuning yields similar minimal generalization errors (under $\bar {\mathcal{P}}_T$ and $\bar {\mathcal{P}}_E$, respectively) as follows:

\begin{equation} \label{close}
\text{\scriptsize $\left| \left(\hat{\mathcal{E}}({\mathcal{D}_{n_T}^T},\bar {\mathcal{P}}_T)+O(\sqrt{\frac{|\bar {\mathcal{P}}_T|}{n_T}})\right)- \left(\hat{\mathcal{E}}({\mathcal{D}_{n_T}^T},\bar {\mathcal{P}}_E)+O(\sqrt{\frac{|\bar {\mathcal{P}}_E|}{n_T}})\right) \right| <\varepsilon.$}
\end{equation}

And for downstream task $T$, the optimal size is $|\bar{\mathcal{P}}_T|$ as in Cor. \ref{col_sup}. But Cor. \ref{col_pre} gives $|\bar{\mathcal{P}}_S|>|\bar{\mathcal{P}}_T|$. Therefore, error bound $\mathcal{U}(\mathcal{E}_T)$ under $|\bar{\mathcal{P}}_S|$ is larger than that under the optimal $|\bar{\mathcal{P}}_T|$: 
\begin{equation} \label{inequa}
\begin{aligned}
     min(\mathcal{U}(\mathcal{E}_T)) &=\hat{\mathcal{E}}({\mathcal{D}_{n_T}^T},\bar {\mathcal{P}}_T)+O\left(\sqrt{|\bar {\mathcal{P}}_T|/n_T}\right)\\ &< \hat{\mathcal{E}}({\mathcal{D}_{n_T}^T},\bar {\mathcal{P}_S})+O\left(\sqrt{|\bar {\mathcal{P}_S}|/n_T}\right)
\end{aligned}
\end{equation}

This means reducing the size of parameter space leads to a decrease in test error, which is aligned with the phenomena in the classical regime. We also empirically validate this trend in Appendix \tmprefdelta{app_delta}. We define this \textbf{O}verfitting mitigation \textbf{G}ain as: $\mathbf{OG}=\hat{\mathcal{E}}({\mathcal{D}_{n_T}^T},\bar {\mathcal{P}_S})+O(\sqrt{|\bar {\mathcal{P}_S}|/n_T})-(\hat{\mathcal{E}}({\mathcal{D}_{n_T}^T},\bar {\mathcal{P}}_T)+O(\sqrt{|\bar {\mathcal{P}}_T|/n_T}))$.

Combining Equation \ref{equa_fine},\ref{equa_delta}, Inequality \ref{close}, and definition of $\mathbf{OG}$, we obtain the following proposition:

\begin{proposition} \textbf{Conditioned on $\varepsilon<\mathbf{OG}$, PEFT has tighter bounds than fine-tuning.} Compare the tightest upper bound of PEFT on $|\bar {\mathcal{P}_E}|$ (where $|\bar {\mathcal{P}_E}|<|\bar {\mathcal{P}_S}|$) with the bounds of fine-tuning:
\begin{equation}
    \begin{aligned}
     & \mathcal{U}(\mathcal{E}_F)-min(\mathcal{U}(\mathcal{E}_E)) \\
    >& \text{\scriptsize $\hat{\mathcal{E}}({\mathcal{D}_{n_T}^T},\bar {\mathcal{P}_S})+O(\sqrt{\frac{|\bar {\mathcal{P}_S}|}{n_T}}) - \left(\hat{\mathcal{E}}({\mathcal{D}_{n_T}^T},\bar {\mathcal{P}}_T)+O(\sqrt{\frac{|\bar {\mathcal{P}}_T|}{n_T}})\right)-\varepsilon$}\\
    =& \mathbf{OG}-\varepsilon
    \end{aligned}
\end{equation}

If PEFT preserves enough expressivity and $\mathbf{OG}$ is large enough, the condition $\varepsilon<\mathbf{OG}$ is satisfied. Therefore, PEFT provides tighter bounds than full fine-tuning. The best approach is pre-training on a larger model and utilizing PEFT with comparable expressivity and much fewer parameters.
\end{proposition}

\section{Methodology}

\subsection{AdapterGNN}
We propose a GNN PEFT framework called AdapterGNN. The framework is demonstrated in Fig. \ref{fig_framework}. 
It adds trainable adapters in parallel to GNN MLPs, combining task-specific knowledge in adapters with task-agnostic knowledge from the pre-trained model.
The multi-layer perception (MLP) module contains the majority of the learnable parameters and is important for GNNs. Therefore, we introduce adapters as parallel computations to the GNN MLPs. It utilize bottleneck architecture which includes a down-projection $\mathbf{W}_{\rm down}: \mathbb{R}^{n_{\rm in}}\rightarrow \mathbb{R}^{n_{\rm mid}}$, a ReLU activation, and an up-projection $\mathbf{W}_{\rm up}: \mathbb{R}^{n_{\rm mid}}\rightarrow \mathbb{R}^{n_{\rm out}}$. Unlike the original GNN MLP, the middle dimension $n_{\rm mid}$ is greatly reduced (e.g., $20\times$) as a bottleneck, resulting in a significant reduction in the size of tunable parameters space.

AdapterGNN utilizes ideas adopted in adapters \cite{houlsby2019parameter}. However, different from transformer-based models, GNNs possess distinctive characteristics. It needs to aggregate information from the neighborhood. The prerequisite in the justification Section gives that preserving the expressivity of the full fine-tuning GNNs is crucial for PEFT to generalize better than full fine-tuning. To enhance the expressivity in non-transformer-based GNNs, AdapterGNN employs several novel techniques specially designed:

\paragraph{Dual adapter modules.} In each GNN layer, message passing (MP) aggregates information from the neighborhood. Node embeddings before and after MP provides complementary and informative information. To sufficiently capture the complementary information, we adopt dual parallel adapters. The first adapter takes input before MP, able to deal with the original information. And the other adapter takes input after the non-parametric MP can effectively adapt the MP-aggregated information.

\paragraph{Batch normalization (BN).} Vanilla adapters did not utilize BN. However, data distribution may shift in GNNs. It is essential to ensure that affine parameters in BN are tunable during training. To maintain consistency with the backbone output, we incorporate tunable BN within each adapter.

\paragraph{Learnable scaling.} 
The vanilla adapter simply adds the output of an adapter to the original output without scaling. This often causes catastrophic forgetting. On the other hand, manually tuning \cite{hu2021lora, chen2022adaptformer} the scaling factors requires extensive effort and time, as the optimal scaling factors may vary significantly across different datasets. This goes against our objective of achieving efficient training. In AdapterGNN, we use learnable scaling factors $s_1, s_2$ with a small start value. They can be dynamically adjusted during training.

Our experiments demonstrate that AdapterGNN can achieve stable and superior performance with these novel techniques.
Formally, the output of the adapter is:
\begin{equation}
    \mathbf{A}(\mathbf{x})={\rm BN}(\mathbf{W}_{\rm up}({\rm ReLU}(\mathbf{W}_{\rm down}(\mathbf{x})))).
\end{equation}
The output of adapters and the original embedding are combined by element-wise addition:
\begin{equation}
    \mathbf{h}_{l} = {\rm BN}({\rm MLP}({\rm MP}(\mathbf{x}_l))) + s_1\cdot \mathbf{A}_1(\mathbf{x}_l) + s_2\cdot \mathbf{A}_2({\rm MP}(\mathbf{x}_l)).
\end{equation}
where $\mathbf{x}_l$ is the input of the $l$th layer and $\mathbf{h}_{l}$ is the embedding before GNN $\rm ReLU$ and $\rm 
 Dropout$, i.e., $\mathbf{x}_{l+1}={\rm Dropout}({\rm ReLU}(\mathbf{h}_l))$.

\subsection{Discussion}
Our framework provides several advantages. Firstly, the issue of catastrophic forgetting of pre-trained knowledge, is effectively mitigated through our parallel adapter design and novel learnable scaling strategy. This strategy automatically controls the proportion of newly-tuned task-specific knowledge, while persistently preserving the pre-trained knowledge. Secondly, to address overfitting when tuning on a small dataset with large parameters. Our framework introduces a bottleneck that significantly reduces the size of tunable parameter space to alleviate overfitting and improve generalization, while also being highly efficient. Additionally, AdapterGNN is specifically designed for GNNs and takes input both before and after message passing. This dual structure along with BN maximally preserves expressivity to satisfy the prerequisite to fully exploit the benefits of PEFT.
Although our architecture is specifically designed for GIN \cite{xu2018powerful}, it can be transferred to any existing GNN architecture such as GAT \cite{velickovic2017graph} and GraphSAGE \cite{hamilton2017inductive}.

\section{Experiments}

\paragraph{Experimental setup.}
We evaluate the effectiveness of AdapterGNN by conducting extensive graph-level classification experiments on eight molecular datasets and one biology dataset. We employ prevalent pre-training methods, all based on a GIN backbone. Details of datasets and pre-trained models are in Appendix \tmprefdataset{app_dataset} and \tmprefpretrain{app_pretrain}. Implementations can be found in Appendix \tmprefdetail{app_imple_detail}.

\begin{table*}[h!]
    \caption{Test ROC-AUC (\%) performances on molecular prediction benchmarks with different tuning methods and pre-trained GNN models.}
    \label{table_main}
    \centering
    \scriptsize
    \begin{tabular}{ccp{0.26in}p{0.26in}p{0.26in}p{0.26in}p{0.3in}p{0.26in}p{0.26in}p{0.26in}|c}
    \toprule
Tuning & Pre-training & \multicolumn{8}{c}{\small{Dataset}} & ~ \\
        Method & Method & \hfil BACE & \hfil BBBP & ClinTox & \hfil HIV & \hfil SIDER & \hfil Tox21 & \hfil MUV & ToxCast & \hfil Avg. \\ \midrule
        \multirow{5}{*}{\shortstack{Full Fine-tune\\(100\%)}} & EdgePred & 79.9\tiny{\textcolor{gray}{±0.9}} 
        & 67.3\tiny{\textcolor{gray}{±2.4}} & 64.1\tiny{\textcolor{gray}{±3.7}} & 76.3\tiny{\textcolor{gray}{±1.0}} & 60.4\tiny{\textcolor{gray}{±0.7}} & \textbf{76.0}\tiny{\textcolor{gray}{±0.6}} & 74.1\tiny{\textcolor{gray}{±2.1}} & 64.1\tiny{\textcolor{gray}{±0.6}} & 70.3 \\
        ~ & ContextPred & \textbf{79.6}\tiny{\textcolor{gray}{±1.2}} & 68.0\tiny{\textcolor{gray}{±2.0}} & 65.9\tiny{\textcolor{gray}{±3.8}} & \textbf{77.3}\tiny{\textcolor{gray}{±1.0}} & 60.9\tiny{\textcolor{gray}{±0.6}} & \textbf{75.7}\tiny{\textcolor{gray}{±0.7}} & 75.8\tiny{\textcolor{gray}{±1.7}} & \textbf{63.9}\tiny{\textcolor{gray}{±0.6}} & 70.9 \\
        ~ & AttrMasking & 79.3\tiny{\textcolor{gray}{±1.6}} & 64.3\tiny{\textcolor{gray}{±2.8}} & 71.8\tiny{\textcolor{gray}{±4.1}} & \textbf{77.2}\tiny{\textcolor{gray}{±1.1}} & 61.0\tiny{\textcolor{gray}{±0.7}} & \textbf{76.7}\tiny{\textcolor{gray}{±0.4}} & 74.7\tiny{\textcolor{gray}{±1.4}} & \textbf{64.2}\tiny{\textcolor{gray}{±0.5}} & 71.1 \\
        ~ & GraphCL & 74.6\tiny{\textcolor{gray}{±2.2}} & 68.6\tiny{\textcolor{gray}{±2.3}} & 69.8\tiny{\textcolor{gray}{±7.2}} & 78.5\tiny{\textcolor{gray}{±1.2}} & 59.6\tiny{\textcolor{gray}{±0.7}} & 74.4\tiny{\textcolor{gray}{±0.5}} & 73.7\tiny{\textcolor{gray}{±2.7}} & 62.9\tiny{\textcolor{gray}{±0.4}} & 70.3 \\
        ~ & SimGRACE & 74.7\tiny{\textcolor{gray}{±1.0}} & \textbf{69.0}\tiny{\textcolor{gray}{±1.0}} & 59.9\tiny{\textcolor{gray}{±2.3}} & 74.6\tiny{\textcolor{gray}{±1.2}} & 59.1\tiny{\textcolor{gray}{±0.6}} & 73.9\tiny{\textcolor{gray}{±0.4}} & 71.0\tiny{\textcolor{gray}{±1.9}} & 61.8\tiny{\textcolor{gray}{±0.4}} & 68.0 \\
        \hline
        \hline

        \multirow{5}{*}{\shortstack{Adapter \\ (5.2\%)}} & 
        EdgePred & 78.5\tiny{\textcolor{gray}{±1.7}} & 65.9\tiny{\textcolor{gray}{±2.8}} & 66.6\tiny{\textcolor{gray}{±5.4}} & 73.5\tiny{\textcolor{gray}{±0.2}} & 60.9\tiny{\textcolor{gray}{±1.3}} & 75.4\tiny{\textcolor{gray}{±0.5}} & 73.0\tiny{\textcolor{gray}{±1.0}} & 63.0\tiny{\textcolor{gray}{±0.7}} & 69.6 \\
        ~ & ContextPred & 75.0\tiny{\textcolor{gray}{±3.3}} & 68.2\tiny{\textcolor{gray}{±3.0}} & 57.6\tiny{\textcolor{gray}{±3.6}} & 75.4\tiny{\textcolor{gray}{±0.6}} & 62.4\tiny{\textcolor{gray}{±1.2}} & 74.7\tiny{\textcolor{gray}{±0.7}} & 73.3\tiny{\textcolor{gray}{±0.8}} & 62.2\tiny{\textcolor{gray}{±0.4}} & 68.6 \\
        ~ & AttrMasking & 76.1\tiny{\textcolor{gray}{±1.4}} & 68.7\tiny{\textcolor{gray}{±1.7}} & 65.8\tiny{\textcolor{gray}{±4.4}} & 75.6\tiny{\textcolor{gray}{±0.7}} & 59.8\tiny{\textcolor{gray}{±1.7}} & 74.4\tiny{\textcolor{gray}{±0.9}} & 75.8\tiny{\textcolor{gray}{±2.4}} & 62.6\tiny{\textcolor{gray}{±0.8}} & 69.9 \\
        ~ & GraphCL & 72.5\tiny{\textcolor{gray}{±3.0}} & \textbf{69.3}\tiny{\textcolor{gray}{±0.6}} & 67.3\tiny{\textcolor{gray}{±7.4}} & 75.0\tiny{\textcolor{gray}{±0.4}} & \textbf{59.7}\tiny{\textcolor{gray}{±1.2}} & 74.7\tiny{\textcolor{gray}{±0.4}} & 72.9\tiny{\textcolor{gray}{±1.7}} & 62.9\tiny{\textcolor{gray}{±0.4}} & 69.3 \\ 
        ~ & SimGRACE & 73.4\tiny{\textcolor{gray}{±1.1}} & 64.8\tiny{\textcolor{gray}{±0.7}} & 63.5\tiny{\textcolor{gray}{±4.4}} & 73.9\tiny{\textcolor{gray}{±1.0}} & \textbf{59.9}\tiny{\textcolor{gray}{±0.9}} & 73.1\tiny{\textcolor{gray}{±0.9}} & 70.1\tiny{\textcolor{gray}{±4.6}} & 61.7\tiny{\textcolor{gray}{±0.8}} & 67.6 \\
        \hline

        \multirow{5}{*}{\shortstack{LoRA \\ (5.0\%)}} & 
        EdgePred & \textbf{81.0}\tiny{\textcolor{gray}{±0.8}} & 64.8\tiny{\textcolor{gray}{±1.6}} & \textbf{67.7}\tiny{\textcolor{gray}{±1.2}} & 74.8\tiny{\textcolor{gray}{±1.2}} & 60.8\tiny{\textcolor{gray}{±1.1}} & 74.6\tiny{\textcolor{gray}{±0.4}} & 75.0\tiny{\textcolor{gray}{±1.5}} & 62.2\tiny{\textcolor{gray}{±1.0}} & 70.1 \\
        ~ & ContextPred & 78.5\tiny{\textcolor{gray}{±1.1}} & 65.3\tiny{\textcolor{gray}{±2.4}} & 61.3\tiny{\textcolor{gray}{±1.9}} & 74.7\tiny{\textcolor{gray}{±1.6}} & 60.8\tiny{\textcolor{gray}{±0.4}} & 72.9\tiny{\textcolor{gray}{±0.4}} & 75.4\tiny{\textcolor{gray}{±0.9}} & 63.4\tiny{\textcolor{gray}{±0.2}} & 69.0 \\
        ~ & AttrMasking & \textbf{79.8}\tiny{\textcolor{gray}{±0.7}} & 64.2\tiny{\textcolor{gray}{±1.1}} & 70.1\tiny{\textcolor{gray}{±2.9}} & 76.1\tiny{\textcolor{gray}{±1.4}} & 59.7\tiny{\textcolor{gray}{±0.5}} & 74.6\tiny{\textcolor{gray}{±0.5}} & 76.6\tiny{\textcolor{gray}{±1.6}} & 61.7\tiny{\textcolor{gray}{±0.4}} & 70.4 \\
        ~ & GraphCL & 75.1\tiny{\textcolor{gray}{±0.7}} & 67.8\tiny{\textcolor{gray}{±1.1}} & 65.1\tiny{\textcolor{gray}{±3.5}} & \textbf{78.9}\tiny{\textcolor{gray}{±0.6}} & 57.6\tiny{\textcolor{gray}{±0.7}} & 73.9\tiny{\textcolor{gray}{±0.9}} & 72.8\tiny{\textcolor{gray}{±1.2}} & 62.7\tiny{\textcolor{gray}{±0.6}} & 69.2 \\ 
        ~ & SimGRACE & 73.2\tiny{\textcolor{gray}{±1.0}} & 67.5\tiny{\textcolor{gray}{±0.4}} & 60.7\tiny{\textcolor{gray}{±0.4}} & 74.1\tiny{\textcolor{gray}{±0.5}} & 57.6\tiny{\textcolor{gray}{±2.6}} & 72.2\tiny{\textcolor{gray}{±0.2}} & 67.9\tiny{\textcolor{gray}{±0.9}} & 61.8\tiny{\textcolor{gray}{±0.2}} & 66.9 \\
        \hline
        \hline
        
        \multirow{5}{*}{\shortstack{GPF \\ (0.1\%)}} & EdgePred & 68.0\tiny{\textcolor{gray}{±0.4}} & 55.9\tiny{\textcolor{gray}{±0.2}} & 50.8\tiny{\textcolor{gray}{±0.1}} & 66.0\tiny{\textcolor{gray}{±0.7}} & 51.5\tiny{\textcolor{gray}{±0.7}} & 63.1\tiny{\textcolor{gray}{±0.5}} & 63.1\tiny{\textcolor{gray}{±0.1}} & 55.7\tiny{\textcolor{gray}{±0.5}} & 59.3 \\
        ~ & ContextPred & 58.7\tiny{\textcolor{gray}{±0.6}} & 58.6\tiny{\textcolor{gray}{±0.6}} & 39.8\tiny{\textcolor{gray}{±0.8}} & 68.0\tiny{\textcolor{gray}{±0.4}} & 59.4\tiny{\textcolor{gray}{±0.2}} & 67.8\tiny{\textcolor{gray}{±0.9}} & 71.8\tiny{\textcolor{gray}{±0.8}} & 58.8\tiny{\textcolor{gray}{±0.5}} & 60.4 \\
        ~ & AttrMasking & 61.7\tiny{\textcolor{gray}{±0.3}} & 54.3\tiny{\textcolor{gray}{±0.3}} & 56.4\tiny{\textcolor{gray}{±0.2}} & 64.0\tiny{\textcolor{gray}{±0.2}} & 52.0\tiny{\textcolor{gray}{±0.2}} & 69.2\tiny{\textcolor{gray}{±0.3}} & 62.9\tiny{\textcolor{gray}{±0.9}} & 58.1\tiny{\textcolor{gray}{±0.3}} & 59.8 \\
        ~ & GraphCL & 71.5\tiny{\textcolor{gray}{±0.6}} & 63.7\tiny{\textcolor{gray}{±0.4}} & 64.5\tiny{\textcolor{gray}{±0.6}} & 70.3\tiny{\textcolor{gray}{±0.5}} & 55.3\tiny{\textcolor{gray}{±0.6}} & 65.5\tiny{\textcolor{gray}{±0.5}} & 70.1\tiny{\textcolor{gray}{±0.7}} & 58.5\tiny{\textcolor{gray}{±0.5}} & 64.9 \\ \hline

        MolCPT (40.0\%) & GraphCL & 74.1\tiny{\textcolor{gray}{±0.5}}  & 60.5\tiny{\textcolor{gray}{±0.8}}  & 73.4\tiny{\textcolor{gray}{±0.8}}  & 64.5\tiny{\textcolor{gray}{±0.8}}  & 55.9\tiny{\textcolor{gray}{±0.3}}  & 67.4\tiny{\textcolor{gray}{±0.7}}  & 65.7\tiny{\textcolor{gray}{±2.2}}  & 57.9\tiny{\textcolor{gray}{±0.3}}  & 64.9 \\  
        \hline
        \hline

        \multirow{5}{*}{\shortstack{\textbf{AdapterGNN} (ours)\\(5.2\%)}} & EdgePred & 79.0\tiny{\textcolor{gray}{±1.5}} & \textbf{69.7}\tiny{\textcolor{gray}{±1.4}} & \textbf{67.7}\tiny{\textcolor{gray}{±3.0}} & \textbf{76.4}\tiny{\textcolor{gray}{±0.7}} & \textbf{61.2}\tiny{\textcolor{gray}{±0.9}} & 75.9\tiny{\textcolor{gray}{±0.9}} & \textbf{75.8}\tiny{\textcolor{gray}{±2.1}} & \textbf{64.2}\tiny{\textcolor{gray}{±0.5}} & \textbf{71.2} \\
        ~ & ContextPred & 78.7\tiny{\textcolor{gray}{±2.0}} & \textbf{68.2}\tiny{\textcolor{gray}{±2.9}} & \textbf{68.7}\tiny{\textcolor{gray}{±5.3}} & 76.1\tiny{\textcolor{gray}{±0.5}} & \textbf{61.1}\tiny{\textcolor{gray}{±1.0}} & 75.4\tiny{\textcolor{gray}{±0.6}} & \textbf{76.3}\tiny{\textcolor{gray}{±1.0}} & 63.2\tiny{\textcolor{gray}{±0.3}} & \textbf{71.0} \\
        ~ & AttrMasking & 79.7\tiny{\textcolor{gray}{±1.3}} & \textbf{67.5}\tiny{\textcolor{gray}{±2.2}} & \textbf{78.3}\tiny{\textcolor{gray}{±2.6}} & 76.7\tiny{\textcolor{gray}{±1.2}} & \textbf{61.3}\tiny{\textcolor{gray}{±1.1}} & 76.6\tiny{\textcolor{gray}{±0.5}} & \textbf{78.4}\tiny{\textcolor{gray}{±0.7}} & 63.6\tiny{\textcolor{gray}{±0.5}} & \textbf{72.8} \\
        ~ & GraphCL & \textbf{76.1}\tiny{\textcolor{gray}{±2.2}} & 67.8\tiny{\textcolor{gray}{±1.4}} & \textbf{72.0}\tiny{\textcolor{gray}{±3.8}} & 77.8\tiny{\textcolor{gray}{±1.3}} & 59.6\tiny{\textcolor{gray}{±1.3}} & \textbf{74.9}\tiny{\textcolor{gray}{±0.9}} & \textbf{75.1}\tiny{\textcolor{gray}{±2.1}} & \textbf{63.1}\tiny{\textcolor{gray}{±0.4}} & \textbf{70.7} \\
        ~ & SimGRACE & \textbf{77.7}\tiny{\textcolor{gray}{±1.7}} & 68.1\tiny{\textcolor{gray}{±1.3}} & \textbf{73.9}\tiny{\textcolor{gray}{±7.0}} & \textbf{75.1}\tiny{\textcolor{gray}{±1.2}} & 58.9\tiny{\textcolor{gray}{±0.9}} & \textbf{74.4}\tiny{\textcolor{gray}{±0.6}} & \textbf{71.8}\tiny{\textcolor{gray}{±1.4}} & \textbf{62.6}\tiny{\textcolor{gray}{±0.6}} & \textbf{70.3} \\
        
        \bottomrule
    \end{tabular}
\end{table*}

\begin{table*}[h!]
  \caption{Test ROC-AUC (\%) performances on PPI benchmark with different tuning methods and pre-trained GNN models. "-" represents no pre-training.}
  \centering
  \small
  \begin{tabular}{ccccccccc}
    \toprule
      Tuning Method & - & \scriptsize{EdgePred} & \scriptsize{ContextPred} & \scriptsize{AttrMasking} & \scriptsize{GraphCL} & \scriptsize{SimGRACE} & Avg.\\
      \midrule

    Full Fine-tune (100\%) & 65.2\tiny{\textcolor{gray}{±1.2}} & 65.6\tiny{\textcolor{gray}{±0.9}} & 63.5\tiny{\textcolor{gray}{±1.1}} & 63.2\tiny{\textcolor{gray}{±1.2}} & 65.5\tiny{\textcolor{gray}{±0.8}} & 68.2\tiny{\textcolor{gray}{±1.2}} & 65.2\\

    GPF (0.1\%) & 65.9\tiny{\textcolor{gray}{±1.9}} & 51.2\tiny{\textcolor{gray}{±1.3}} & 67.1\tiny{\textcolor{gray}{±0.6}} & 69.0\tiny{\textcolor{gray}{±0.3}} & 62.3\tiny{\textcolor{gray}{±0.5}} & 50.0\tiny{\textcolor{gray}{±0.9}} & 60.9\\
    
    Adapter (4.0\%) & 65.6\tiny{\textcolor{gray}{±1.1}} & 69.8\tiny{\textcolor{gray}{±0.5}} & 68.2\tiny{\textcolor{gray}{±1.5}} & \textbf{70.9}\tiny{\textcolor{gray}{±1.0}} & 69.0\tiny{\textcolor{gray}{±0.8}} & 64.5\tiny{\textcolor{gray}{±2.0}} & 68.0\\

    LoRA (4.0\%) & 63.0\tiny{\textcolor{gray}{±0.4}} & 68.0\tiny{\textcolor{gray}{±1.0}} & 68.0\tiny{\textcolor{gray}{±1.1}} & 69.2\tiny{\textcolor{gray}{±0.8}} & \textbf{69.4}\tiny{\textcolor{gray}{±0.6}} & 63.0\tiny{\textcolor{gray}{±0.3}} & 66.8\\

    \textbf{AdpaterGNN} (4.0\%) & \textbf{66.3}\tiny{\textcolor{gray}{±0.9}} & \textbf{70.6}\tiny{\textcolor{gray}{±1.1}} & \textbf{68.3}\tiny{\textcolor{gray}{±1.5}} & 69.7\tiny{\textcolor{gray}{±1.1}} & 68.1\tiny{\textcolor{gray}{±1.5}} & \textbf{70.1}\tiny{\textcolor{gray}{±1.2}} & \textbf{68.9}\\
    \bottomrule
  \end{tabular}
  \label{table_ppi}
\end{table*}

\paragraph{Baseline methods.}
We mainly compare our method with full fine-tuning.
Besides, we make special variations for PEFT methods and implement them in GNNs. Comprehensive comparisons of them are presented in Fig. \ref{fig_compare}, measured by the average ROC-AUC over six small molecular datasets and an AttrMasking pre-trained model. AdapterGNN is the only one surpassing full fine-tuning. Implementations of these PEFT methods in GNNs are in Appendix \tmprefimplementation{app_implementation}. 
We also compare existing PEFT methods in GNNs. GPF \cite{fang2022prompt} only modifies the input by adding a prompt feature to the input embeddings. MolCPT \cite{diao2022molcpt} (backbone frozen) leverages molecular motifs to provide additional information to graph embedding. We omit other related works such as GPPT \cite{sun2022gppt} and GraphPrompt \cite{liu2023graphprompt} because they are either not efficient or cannot yield satisfying performance without a few-shot setting.

\subsection{Main Results}
\label{subsec_exp_main}
We \textbf{bold} the higher one in Table \ref{table_main} and Table \ref{table_ppi} and the results suggests the following:

\textbf{(1) AdapterGNN is the only method consistently outperforming full fine-tuning.} 
On molecular benchmarks, among eight datasets with five pre-training methods, AdapterGNN achieves higher average performances in all pre-training methods. It is the only method that outperforms full fine-tuning. The overall average ROC-AUC is 71.2\%, which is 1.6\% relatively higher than 70.1\% of full fine-tuning. 
On the PPI benchmark, AdapterGNN is the only one that consistently outperforms full fine-tuning. The overall average ROC-AUC is 68.9\%, which is 5.7\% higher.
In detail, AdapterGNN achieves higher performances in 32/46 (70\%) experiments. 
It achieves these with only 5.2\% and 4.0\% tunable parameters, respectively. It also achieves training efficiency (both FLOPs and latency) as detailed in Appendix \tmprefflops{app_flops}.

\textbf{(2) AdapterGNN outperforms state-of-the-art PEFT methods.}
In addition to comparing AdapterGNN with full fine-tuning, we implement variants of parallel adapter \cite{he2021towards} and LoRA \cite{hu2021lora} in GNNs for the first time, which are prevalent PEFT methods in transformer-based models. They are compared under similar proportions of tuned parameters in detail. AdaptGNN has demonstrated consistent and conspicuous improvements on molecular benchmarks. On the PPI benchmark, overfitting is severe for full fine-tuning. Our implemented PEFT methods all achieve superior performance than it. Among them, ApdaterGNN is the best with the highest average.
When compared with existing GNN PEFT methods, the advantage is more significant. Though GPF has a parameter efficiency of only 0.1\%, its performance is limited. And MolCPT tunes much more parameters. On molecular benchmarks, AdapterGNN outperforms GPF and MolCPT by an average of 8.9\% in GraphCL. And on the PPI benchmark, compared with GPF, AdapterGNN outperforms it by an average of 13.1\%.


\textbf{(3) AdapterGNN avoids negative transfer and retains the stable improvements of pre-training.} On the PPI benchmark, negative transfer frequently occurs in full fine-tuning and GPF, where the pre-trained model often yields inferior performance. On the contrary, the benefits of pre-training remain stable in AdapterGNN.

\subsection{Ablation Studies}
\label{subsec_exp_abl}
We investigate the properties that make for a good AdapterGNN. And by ablating on model size and training data size, we validate our theoretical justification for the generalization bounds. All models are pre-trained by AttrMasking. Unless otherwise specified, the performance represents the average ROC-AUC over six small molecular datasets.

\paragraph{Insertion form and BN.} AdapterGNN includes dual adapters parallel to GNN MLP taking input before and after the message passing. To demonstrate the effectiveness of this design, we compare its performance with those of a single parallel adapter and a sequential adapter inserted after GNN MLP. We also compare with the variant without batch normalization (BN). The count of adapter parameters is the same across all forms. Table \ref{table_insert} shows that adopting a single adapter already achieves superior performance over full fine-tuning. Combining two parallel adapters further improves the expressivity of PEFT, yielding the best performance. But without BN, performance drops by a large margin.

\begin{table}
\centering
\caption{Comparison of insertion forms and BN.}
\begin{tabular}{cccc}
    \toprule
      Form & Position & BN & Avg.  \\
    \midrule
    \multicolumn{3}{c}{Full Fine-tune} & 69.6   \\
    Sequential & After MLP & $\checkmark$ &70.2   \\
    Parallel & Before MP & $\checkmark$ & 70.4   \\
    Parallel & After MP & $\checkmark$ &70.3   \\
    Parallel & Dual & $\times$ & 68.0   \\
    Parallel & Dual & $\checkmark$ &\textbf{71.2}   \\
    \bottomrule
  \end{tabular}
  \label{table_insert}
\end{table}

\begin{table}
  \caption{Comparison of learnable scaling and fixed scaling.}
  \centering
  \scriptsize
  \begin{tabular}{p{0.3in}p{0.24in}p{0.24in}p{0.24in}p{0.24in}p{0.24in}p{0.24in}c}
    \toprule
      ~ & BACE & BBBP & ClinTox & SIDER & Tox21 & ToxCast & Avg. \\
    \midrule
    0.01 & 77.8±1.9&	67.6±1.4&	76.3±2.8	&59.9±1.1&	74.8±0.4	&62.5±0.4 & 69.8  \\
    0.1 & 78.5±1.1&	67.6±2.1&	72.6±7.0&	61.0±0.7	&76.2±0.5&	63.3±0.3 & 69.9 \\
    0.5 & 78.7±1.1	&67.6±3.2	&72.3±6.0&	60.9±1.0&	76.2±0.7&	63.3±0.3  & 69.8 \\
    1 & 78.6±1.7	&\textbf{68.7±3.0}&	66.3±7.2&	\textbf{61.3}±0.8 &	75.7±0.7&	63.3±0.6 & 69.0 \\
    5 & 73.7±2.4&	66.6±2.1&	55.9±6.6&	60.8±1.5&	75.3±0.8&	62.9±0.5 & 65.9 \\
    \textbf{Learnable} & \textbf{79.7}±1.3 & 67.5±2.2 & \textbf{78.3}±2.6 & \textbf{61.3}±1.1 & \textbf{76.6}±0.5 & \textbf{63.6}±0.5 & \textbf{71.2}  \\
    \bottomrule
  \end{tabular}
  \label{table_scaling}
\end{table}

\paragraph{Scaling strategy.} We compared our novel learnable scaling strategy with various fixed scaling ranging from 0.01 to 5. Table \ref{table_scaling} shows that in five out of six datasets, as well as on average, our learnable scaling strategy achieved the highest performance. Among the fixed scalings, smaller scaling is superior. As the scaling factor increased, the performance deteriorated due to catastrophic forgetting of pre-trained knowledge.

\begin{figure}[ht]
  \centering
  \begin{minipage}[t]{0.48\linewidth}
    \centering
    \includegraphics[width=\textwidth]{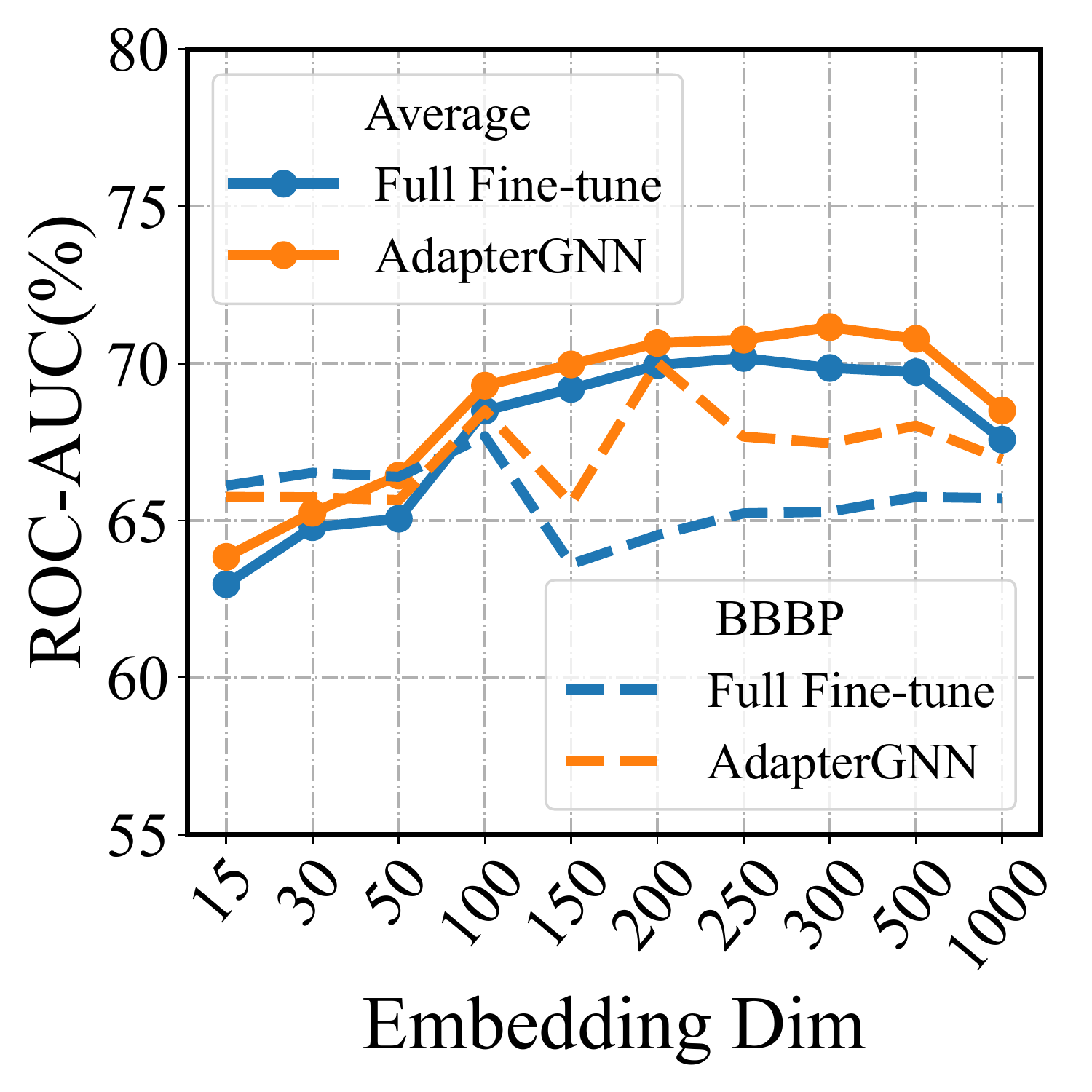}
    {(a) Average (solid) and BBBP performance (dotted).}
  \end{minipage}
  \hfill
  \begin{minipage}[t]{0.48\linewidth}
    \centering
    \includegraphics[width=\textwidth]{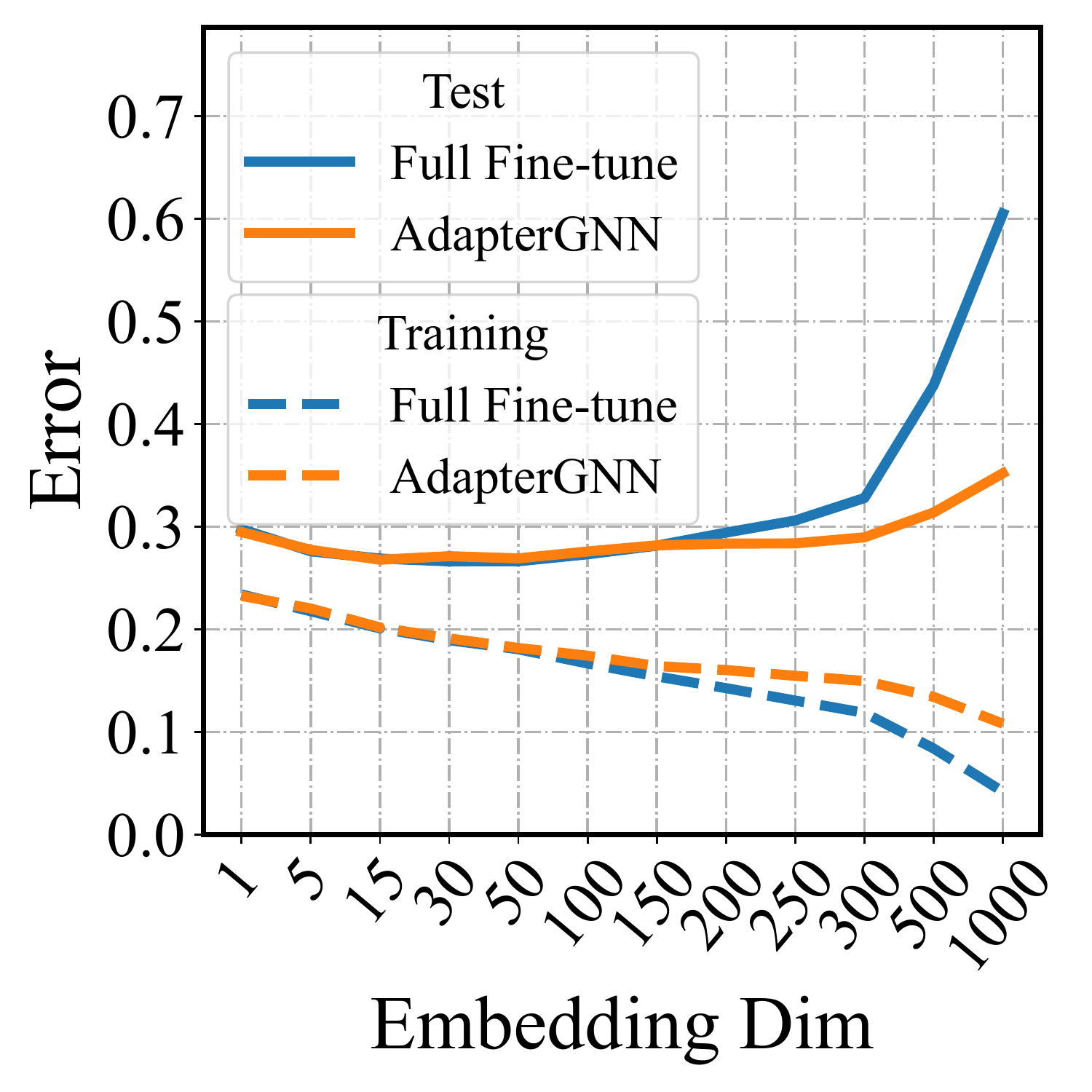}
    {(b) The generalization gap in Tox21.}
  \end{minipage}
  \caption{Test ROC-AUC (\%) performances \& generalization gap with different model sizes.}
  \label{fig_abl_1}
\end{figure}

\paragraph{Model size.} 
We report performances and errors across model sizes by varying embedding dimensions. 
\textbf{(1)} As shown in Fig. \ref{fig_abl_1}(a), average performance initially increases and then decreases, indicating a larger model can be worse, which is consistent with the classical regime. AdapterGNN consistently outperforms full fine-tuning across all model sizes. Specifically, in the BBBP dataset, AdapterGNN may not surpass full fine-tuning with small model sizes, but it achieves superior performance in larger models.
\textbf{(2)} Fig. \ref{fig_abl_1}(b) displays the errors in the Tox21 dataset. The U-shaped test curve also validates the classical regime in our tasks. Although increasing the model size leads to a significant increase in the generalization gap in full fine-tuning, the gap is well-controlled with AdapterGNN. It demonstrates AdapterGNN's superior generalization ability, especially in larger models.

\begin{figure}[ht]
  \centering
  \begin{minipage}[t]{0.48\linewidth}
    \centering
    \includegraphics[width=\textwidth]{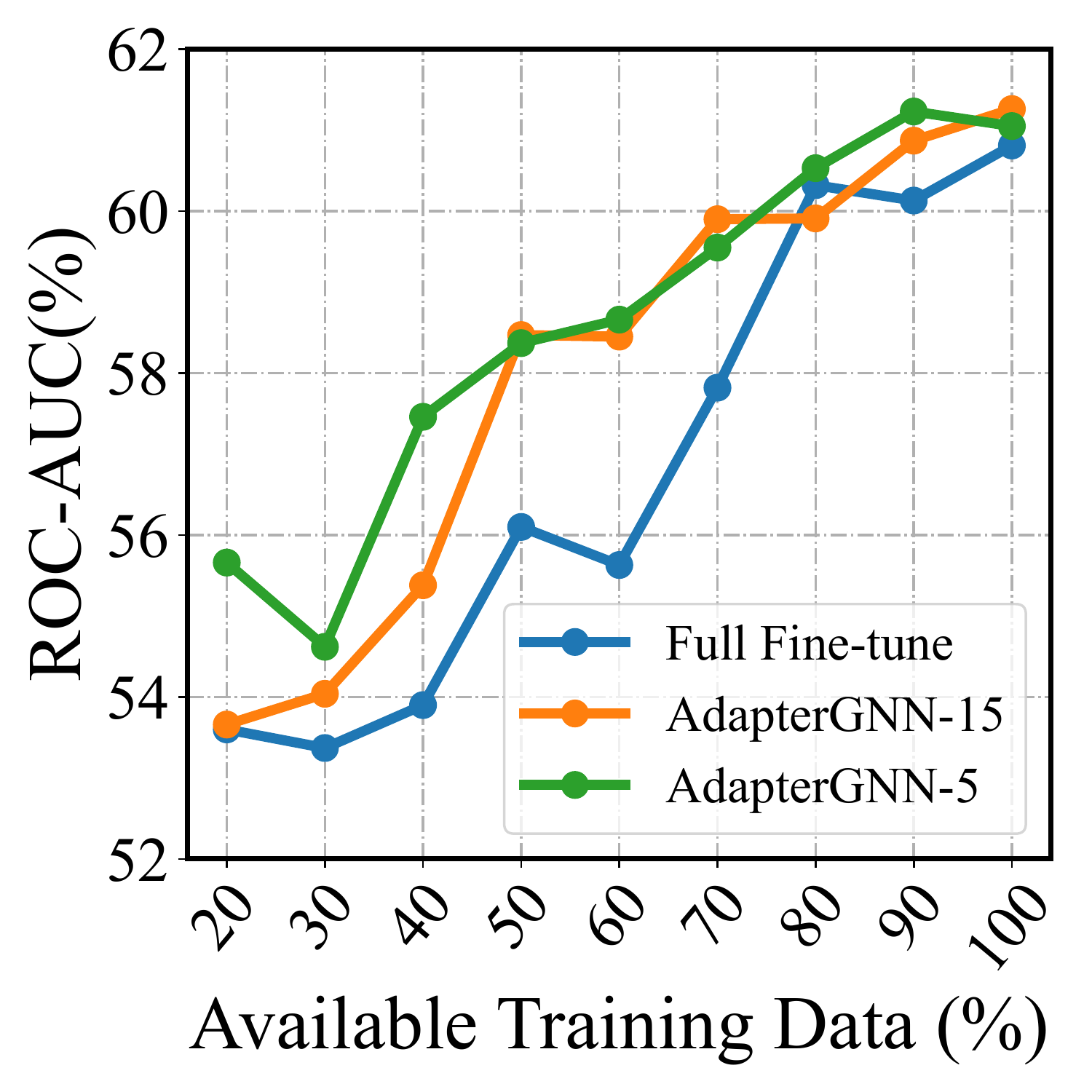}
    {(a) SIDER (1142 training samples).}
  \end{minipage}
  \hfill
  \begin{minipage}[t]{0.48\linewidth}
    \centering
    \includegraphics[width=\textwidth]{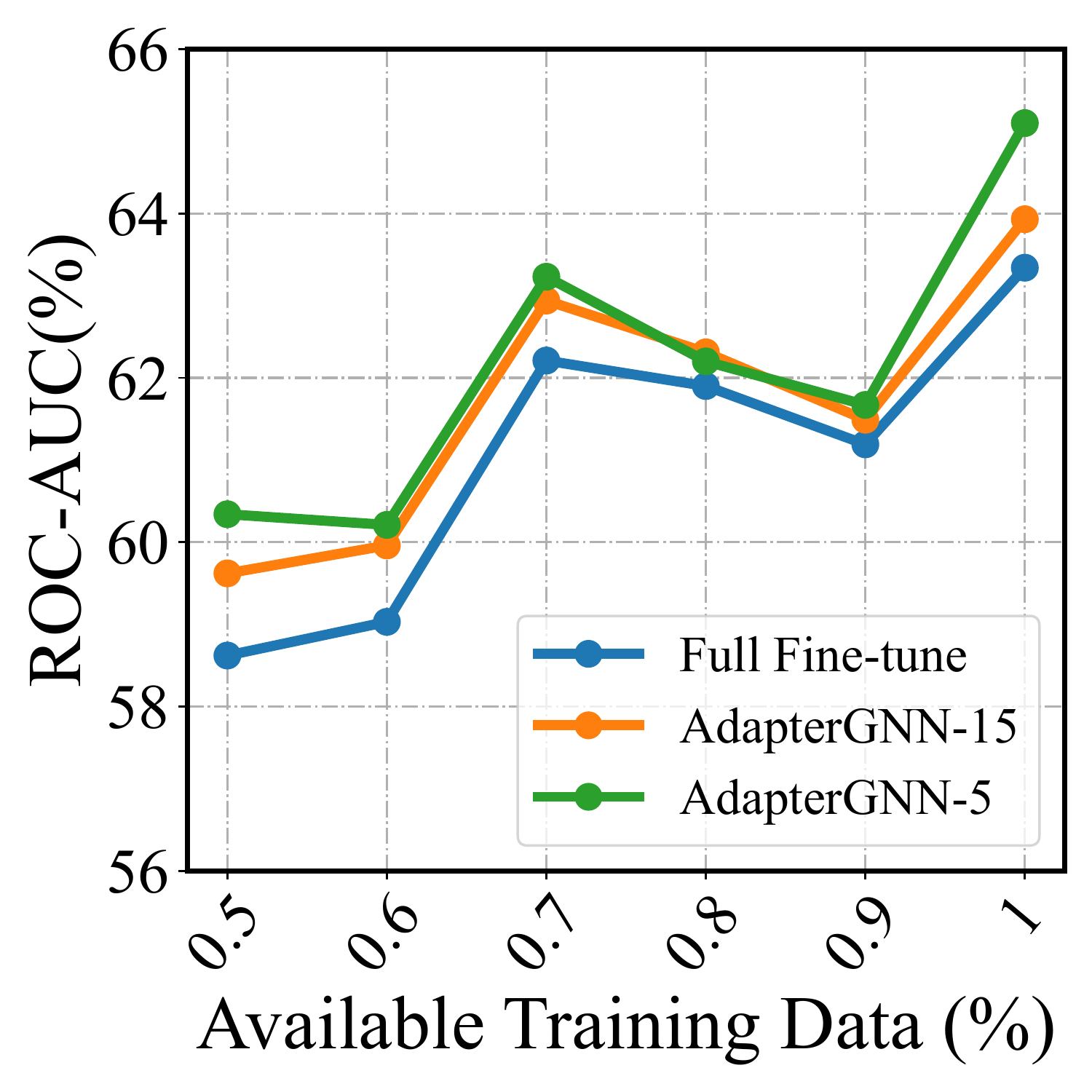}
    {(b) Tox21 (6265 training samples).}
  \end{minipage}
  \caption{Test ROC-AUC (\%) performances with different training data sizes.}
  \label{fig_abl_2}
\end{figure}

\paragraph{Training data size.} Reducing the size of the training samples results in inferior generalization, while AdapterGNN can mitigate this overfitting issue. We compare the performance of full fine-tuning with two AdapterGNN settings, with bottleneck dimensions of 15 (default) and 5, respectively.
Fig. \ref{fig_abl_2} demonstrates that when data becomes scarce, the performance of AdapterGNN with fewer tunable parameters decreases slower and obtains superior results. And fewer parameters yield better results.

\begin{figure}[h!]
\centering
\includegraphics[width=0.45\textwidth]{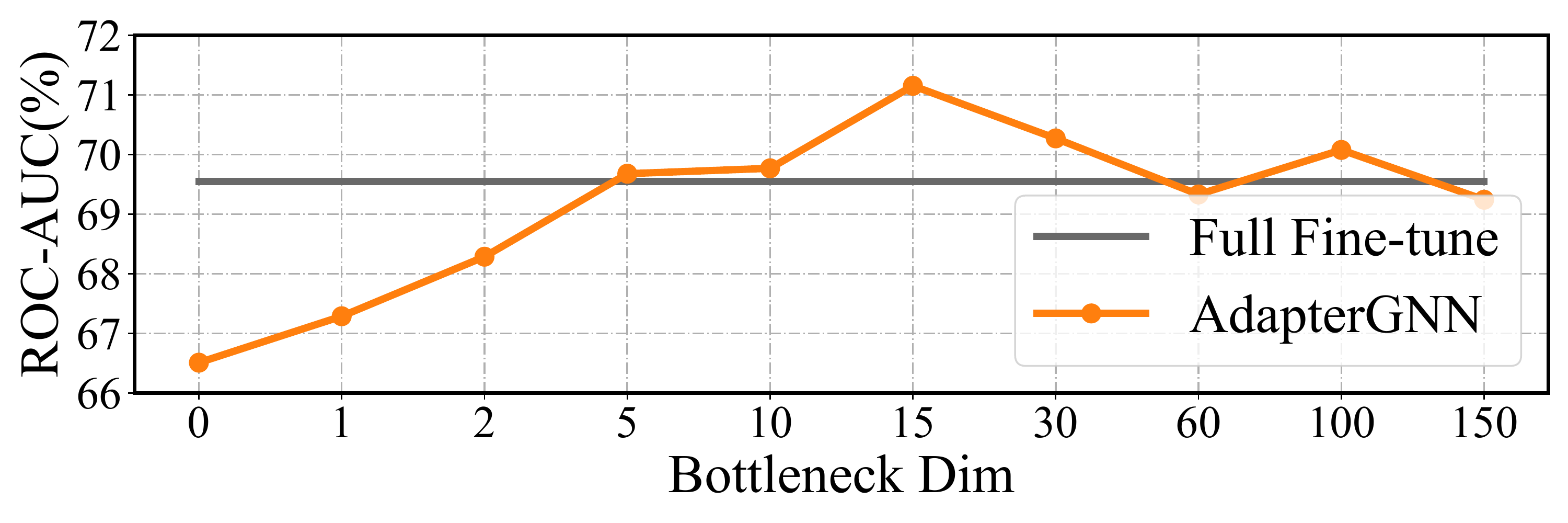}
\caption{Performances with different bottleneck dimensions. 0 represents identical mapping.}
\label{fig_bottleneck}
\end{figure}

\paragraph{Bottleneck dimension.} Fig.\ref{fig_bottleneck} demonstrates that reducing the bottleneck dimension to limit the size of tunable parameter space can improve the generalization ability of the model. But when the size is too small, the model may suffer from underfitting, which can restrict its performance. Therefore, selecting a bottleneck dimension of 15, which represents 5.2\% of all parameters, yields the best average performance. Meanwhile, a dimension of 5, which accounts for only 2.2\% of all parameters, can surpass the results of full fine-tuning.

\section{Conclusion}
We present an effective PEFT framework specially designed for GNNs called AdapterGNN. It is the only PEFT method that outperforms full fine-tuning with much fewer tunable parameters, improving both efficiency and effectiveness. We provide a theoretical justification for this improvement and find that our tasks fall within the classical regime of the generalization error, where larger GNNs can be worse, and reducing the size of the parameter space during tuning can result in lower test error and superior performance. We focus on graph-level classification tasks on GINs and leave other tasks and GNN models for future exploration.

\bibliography{ref}

\appendix

\section{A\ \ \ Preliminaries}
\label{app_preliminary}

Our paper focuses on the graph-level classification task. Let tuple $\mathcal{G}=(\mathcal{V}, \mathcal{E}, \mathbf{X}_{\mathcal{V}}, \mathbf{X}_{\mathcal{E}}) \in \mathbb{G}$ denotes a graph from a dataset $\mathbb{G}$, where $\mathcal{V}=\{v_1,v_2,\cdots v_N\}$ and $\mathcal{E}$ are sets of nodes and edges. The adjacency matrix is $\mathbf{A} \in \{0,1\}^{N\times N}$, where $\mathbf{A}_{ij}=1$ represents $(v_i,v_j)\in \mathcal{E}$. $\mathbf{X}_{\mathcal{V}}\in \mathbb{R}^{N\times \mathbb{F}_{\mathcal{V}}}$ is the node feature matrix, where $x_i\in \mathbb{R}^{\mathbb{F}_{\mathcal{V}}}$ is the feature of the node $v_i$. 
$\mathbf{X}_{\mathcal{E}}\in \mathbb{R}^{M\times \mathbb{F}_{\mathcal{E}}}$ is the edge feature matrix, where $x_{ij}\in \mathbb{R}^{\mathbb{F}_{\mathcal{E}}}$ is the feature of the edge $(v_i,v_j)$. These feature matrices are optional.

\paragraph{Graph-level Classification Task.} A GNN encoder is often represented as $f_{\theta}$ and encodes a graph as $h=f_{\theta}(\mathcal{G})$, where $h$ is the feature emdedding of the input graph $\mathcal{G}$ and $\theta$ is parameters of the encoder.
Then a classifier $g_\phi$ takes $h$ as input and gives classification probability $\hat{y}=g_\phi (h) \in \mathbb{R}^{C}$, where $C$ is the number of classes and $\phi$ is the parameters of the classifier. The predicted output is $argmax(\hat{y})$. And for training, the target loss function is $\mathcal{L}(y, \hat{y})$, where $y$ is the ground truth label for $\mathcal{G}$.

\paragraph{Pre-training and Fine-tuning GNN Models.} The pre-training procedure aims to give a superior initialization of $\theta$, as $\theta_{S}$. It always utilizes larger datasets $\mathbb{G}_{S}$. Here we give the formulation under a supervised manner to pre-train $f_{\theta}$ as an example:

\begin{equation}
\addtocounter{equation}{+8}
    \min_{\theta, \phi}\sum_{(\mathcal{G}_i, y_i) \in \mathbb{G}_{S}}\mathcal{L}^{S}\left(g_{\phi}(f_{\theta}(\mathcal{G}_i)), y_i\right).
\end{equation}

where the optimized $\theta$ is the pre-trained output $\theta_{pre}$. Note that many unsupervised manners \cite{hu2019strategies} are also applicable and the formulation can be diverse.

The pre-trained GNN encoder is task-agnostic and can be transferred to different downstream tasks. However, the classifier is task-specific, so it is not retained. The fine-tuning task is always supervised, allowing us to use the following formulation:

\begin{equation}
\begin{array}{cl}
    \min_{\theta, \phi}\sum_{(\mathcal{G}_i, y_i) \in \mathbb{G}_{T}}\mathcal{L}^{T}\left(g_{\phi}(f_{\theta}(\mathcal{G}_i)), y_i\right), \\
    s.t.\ \ \ \ \theta_{init}=\theta_{S}.
    \end{array}
\end{equation}

\paragraph{PEFT GNN Models.} PEFT is a technique that, similar to fine-tuning, makes use of the pre-trained encoder $\theta_{S}$. However, during the tuning stage, PEFT differs from fine-tuning in that it only updates part of $\theta$ \cite{zaken2021bitfit} or inserts additional modules to $\theta$ and updates only the inserted modules \cite{houlsby2019parameter, li2021prefix, hu2021lora, lester2021power}. In the context of GNNs, these parameters can be represented as $\psi$, and they are updated in conjunction with the classifier:

\begin{equation}
    \min_{\psi, \phi}\sum_{(\mathcal{G}_i, y_i) \in \mathbb{G}_{T}}\mathcal{L}^{T}\left(g_{\phi}(f_{\theta_{S},\psi}(\mathcal{G}_i)), y_i\right).
\end{equation}

where $f_{\theta_{S},\psi}$ is the modified encoder and the original parameters $\theta_{S}$ (or most of them) are fixed during PEFT. PEFT is more efficient and the tunable parameter space is much smaller than the original: $ |\psi|\ll |\theta|$.

\section{B\ \ \ Detailed Derivations}

\subsection{B.1 Detailed Proof for Generalization Bounds for Finite Hypothesis Space in Classical Regime}
\label{app_finite}
\begin{proof}
Let $\mathcal{H}$ be a finite hypothesis space, $h \in \mathcal{H}$ is a group of trained parameters within parameter space $\mathcal{H}$. $\hat{\mathcal{E}}_{\mathcal{D}_n}$ represents training error over sampled training data $\mathcal{D}_n$, and $\mathcal{E}$ represents test error. $n$ is the number of training data.

We begin by analyzing the probability $\delta$ that the test error for a hypothesis space $h$ will surpass the training error by a value greater than $\varepsilon$:
\begin{equation}
    \delta = P\left(\exists h \in \mathcal{H},\left|\hat{\mathcal{E}}_{\mathcal{D}_{n}}(h)-\mathcal{E}(h)\right| \geq \varepsilon\right).
    \label{delta}
\end{equation}

From the theorem of the union bound of probability, we can get:

\begin{equation}
    \begin{aligned}
        \delta &\text{\scriptsize $= P\left(\exists h \in \mathcal{H},\left|\hat{\mathcal{E}}_{\mathcal{D}_{n}}(h)-\mathcal{E}(h)\right| \geq \varepsilon\right)$} \\ &\text{\scriptsize $=P\left(\left[\left|\hat{\mathcal{E}}_{\mathcal{D}_{n}}\left(h_{1}\right)-\mathcal{E}\left(h_{1}\right)\right| \geq \varepsilon\right] \vee \cdots \vee\left[\left|\hat{\mathcal{E}}_{\mathcal{D}_{n}}\left(h_{|\mathcal{H}|}\right)-\mathcal{E}\left(h_{|\mathcal{H}|}\right)\right| \geq \varepsilon\right]\right)$}\\
        &\text{\scriptsize $\leq \sum_{h \in \mathcal{H}}P\left(\left|\hat{\mathcal{E}}_{\mathcal{D}_{n}}(h)-\mathcal{E}(h)\right| \geq \varepsilon\right)$}.
    \end{aligned}
    \label{union}
\end{equation}

Then we refer to Hoeffding’s Inequality:
\begin{theorem}
    Hoeffding’s Inequality: Let $\{X_i\} =\{X_1, X_2, \cdots, X_n\}$ be independent random variables such that $a_i\leq X_i\leq b_i$ almost surely. The following inequality holds:
    \begin{equation}
        P\left(\left|\sum_{i=1}^{n}\left(X_{i}-\mathbb{E} X_{i}\right)\right| \geq \varepsilon\right) \leq 2 \exp \left[-\frac{2 \varepsilon^{2}}{\sum_{i=1}^{n}\left(b_{i}-a_{i}\right)^{2}}\right].
        \label{hoeffding}
    \end{equation}
\end{theorem}

We define our training error as $X_{i}=\ell\left(h\left(\mathbf{x}_{i}\right), y_{i}\right)$ for $(\mathbf{x}_{i},y_i)\in \mathcal{D}_{n}$. Thus the test error is the expectation of $X_{i}$. We can give:

\begin{equation}
\begin{aligned}
    \text{\scriptsize $\sum_{i=1}^{n}\left(X_{i}-\mathbb{E} X_{i}\right)$} &\text{\scriptsize $=n\left\{\left[\frac{1}{n} \sum_{i=1}^{n} \ell\left(h\left(\boldsymbol{x}_{i}\right), y_{i}\right)\right]-\mathbb{E}_{(\boldsymbol{x}, y) \sim D} \ell(h(\boldsymbol{x}), y)\right\}$} \\
&\text{\scriptsize $=n\left(\hat{\mathcal{E}}_{\mathcal{D}_{n}}(h)-\mathcal{E}(h)\right).$}
    \end{aligned}
\end{equation}

Meanwhile, we assume that 01-loss is adopted as $\ell\left(h\left(\boldsymbol{x}_{i}\right), y_{i}\right) \in[0,1]=\left[a_{i}, b_{i}\right]$. Therefore, taking the above conditions into Eq. \ref{hoeffding}, we obtain:
\begin{equation}
    P\left(n\left|\hat{\mathcal{E}}_{\mathcal{D}_{n}}(h)-\mathcal{E}(h)\right| \geq \varepsilon\right) \leq 2 e^{-\frac{2 \varepsilon^{2}}{n}}.
\end{equation}

Taking it back to Eq. \ref{union}:

\begin{equation}
    \delta \leq 2|\mathcal{H}|\exp{(-2n\varepsilon^{2})}.
\end{equation}

\begin{equation}
    \log\frac{\delta}{2} \leq \log|\mathcal{H}|-2n\varepsilon^{2}.
\end{equation}

\begin{equation}
    \varepsilon \leq \sqrt{ \frac{\log|\mathcal{H}|-\log\frac{\delta}{2}}{2n}}.
\end{equation}

Note that Eq. \ref{delta} is equivalent to: 
\begin{equation}
        1-\delta = P\left(\forall h \in \mathcal{H},\left|\hat{\mathcal{E}}_{\mathcal{D}_{n}}(h)-\mathcal{E}(h)\right| < \varepsilon\right).
\end{equation}

Finally, we have:
\begin{equation}
    \text{\scriptsize $P\left(\forall h \in \mathcal{H},\left|\hat{\mathcal{E}}_{\mathcal{D}_{n}}(h)-\mathcal{E}(h)\right| \leq \sqrt{ \frac{\log|\mathcal{H}|-\log\frac{\delta}{2}}{2n}} \right) \geq 1-\delta.$}
\end{equation}

Therefore, with probability at least $1-\delta$:

\begin{equation}
    \forall h \in \mathcal{H}, \quad \mathcal{E}(h) \leq \hat{\mathcal{E}}_{\mathcal{D}_n}(h)+O\left(\sqrt{\frac{\log |\mathcal{H}|+\log \frac{2}{\delta}}{2 n}}\right).
\end{equation}

The theorem states that the test error is statistically bounded by three factors: the training error $\hat{\mathcal{E}}$, the size of the hypothesis space $\log |\mathcal{H}|$, and the number of training samples $n$.

For simplicity, we omit the probability notation and use $\mathcal{U}$ to represent the upper bound. We will also omit the $\log$ term and use $\mathcal{P}$ to represent the parameter space. Therefore, $|\mathcal{P}|$ will quantify the size of the parameter space. Sampled data $\mathcal{D}_n$ and the trained parameters $\mathcal{P}$ are variables of training error $\hat{\mathcal{E}}$:

\begin{equation}
    \mathcal{E} \leq\mathcal{U}(\mathcal{E})=\hat{\mathcal{E}}({\mathcal{D}_n},\mathcal{P})+O\left(\sqrt{|\mathcal{P}|/n}\right).
\end{equation}
\end{proof}

\subsection{B.2 Detailed Proof for Corollary \tmprefcolsup{col_sup}}
\label{app_sup}

\begin{proof}
For the upper bound of the task: $\mathcal{U}(\mathcal{E}_T) = \hat{\mathcal{E}}({\mathcal{D}_{n_T}^T},\mathcal{P}_T)+O\left(\sqrt{|\mathcal{P}_T|/n_T}\right)
$. First, we can infer that With the increase of $|\mathcal{P}_T|$, a larger parameter size confers stronger optimization ability, leading to the decrease of the first term training error $\hat{\mathcal{E}}(\mathcal{D}_{n_T}^T, \mathcal{P}_T)$.

Meanwhile, the second term generalization gap bounds $O\left(\sqrt{|\mathcal{P}_T|/n_T}\right)$ increases.

We can get the partial derivative of $\mathcal{U}(\mathcal{E}_T)$ along with  $|\mathcal{P}_T|$:

\begin{equation}
\begin{aligned}
    \frac{\partial(\mathcal{U}(\mathcal{E}_T))}{\partial|\mathcal{P}_T|}=&\frac{\partial(\hat{\mathcal{E}}({\mathcal{D}_{n_T}^T},\mathcal{P}_T))}{\partial|\mathcal{P}_T|}+\frac{\partial O\left(\sqrt{|\mathcal{P}_T|/n_T}\right)}{\partial|\mathcal{P}_T|}\\=&\frac{\partial(\hat{\mathcal{E}}({\mathcal{D}_{n_T}^T},\mathcal{P}_T))}{\partial|\mathcal{P}_T|}+O(1/\sqrt{|\mathcal{P}_T|\cdot n_T}).
\end{aligned}
\end{equation}

The first term $\frac{\partial(\hat{\mathcal{E}}({\mathcal{D}_{n_T}^T},\mathcal{P}_T))}{\partial|\mathcal{P}_T|}<0$ and the second term $O(1/\sqrt{|\mathcal{P}_T|\cdot n_T})>0$. 

Empirically, we have observed that when the size of parameter space, denoted by $|\mathcal{P}_T|$, is extremely large, the risk of overfitting is high, leading to a large test error $\mathcal{U}(\mathcal{E}_T)$. Conversely, when $|\mathcal{P}_T|$ is extremely small, the model may not be able to fit the data well, again resulting in a large test error $\mathcal{U}(\mathcal{E}_T)$.

Therefore, there exists an extreme point where the partial derivative of $\mathcal{U}(\mathcal{E}_T)$ with respect to $|\mathcal{P}_T|$ is zero, and this point corresponds to a minimum. Given a fixed $n_T$ and ${\mathcal{D}^T_{n_T}}$, we can choose the optimal value of $|\bar {\mathcal{P}_T}|$ to minimize the test error $\mathcal{U}(\mathcal{E}_T)$, where $\partial(\hat{\mathcal{E}}({\mathcal{D}_{n_T}^T},\bar {\mathcal{P}_T}))/\partial| \mathcal{P}_T|+O(1/\sqrt{|\bar {\mathcal{P}_T}|\cdot n_T})=0$.
\end{proof}

\subsection{B.3 Detailed Proof for Corollary \tmprefcolpre{col_pre}}
\label{app_pre}

\begin{proof}
Data for the pre-training task $S$ is more abundant than downstream task $T$, satisfying $n_S>n_T$. For task $S$, we have the generalization upper bounds as: $\mathcal{U}(\mathcal{E}_S) = \hat{\mathcal{E}}({\mathcal{D}_{n_S}^S},\mathcal{P}_S)+O\left(\sqrt{|\mathcal{P}_S|/n_S}\right)
$.

From Coll.\tmprefcolsup{col_sup}, we have known that:

\begin{equation}
\text{\scriptsize $\frac{\partial(\mathcal{U}(\mathcal{E}_T))}{\partial|\mathcal{P}|}_{|\mathcal{P}|=|\bar{\mathcal{P}_T}|}=\frac{\partial(\hat{\mathcal{E}}({\mathcal{D}_{n_T}^T},\bar {\mathcal{P}_T}))}{\partial| \mathcal{P}|}+O(1/\sqrt{|\bar {\mathcal{P}_T}|\cdot n_T})=0.$}
\end{equation}

Practically, we often leverage data from the same source to pre-train. For example, unlabeled molecular graphs are used to pre-train and the labeled ones are for fine-tuning in our experiments. The property of data is similar: $ \hat{\mathcal{E}}({\mathcal{D}_{n_T}^T},\mathcal{P})\approx\hat{\mathcal{E}}({\mathcal{D}_{n_S}^S},\mathcal{P})$

Considering $n_S>n_T$ and $O(1/\sqrt{|\mathcal{P}_T|\cdot n_T})>0$, we can obtain the following inequality:

\begin{equation}
\text{\scriptsize $\frac{\partial(\mathcal{U}(\mathcal{E}_S))}{\partial|\mathcal{P}|}_{|\mathcal{P}|=|\bar{\mathcal{P}_T}|}=\frac{\partial(\hat{\mathcal{E}}({\mathcal{D}_{n_S}^S},\bar {\mathcal{P}_T}))}{\partial| \mathcal{P}|}+O(1/\sqrt{|\bar {\mathcal{P}_T}|\cdot n_S})<0. $}
\end{equation}

Meanwhile, for $S$, we have the optimal $|\bar {\mathcal{P}_S}|$ to get:

\begin{equation}
    \text{\scriptsize $\frac{\partial(\mathcal{U}(\mathcal{E}_S))}{\partial|\mathcal{P}|}_{|\mathcal{P}|=|\bar{\mathcal{P}_S}|}=\frac{\partial(\hat{\mathcal{E}}({\mathcal{D}_{n_S}^S},\bar {\mathcal{P_S})})}{\partial| \mathcal{P}|}+O(1/\sqrt{|\bar {\mathcal{P}_S}|\cdot n_S})=0.$}
\end{equation}

Combining the above two formulas and recall  $|\bar {\mathcal{P}_S}|$ is the minimum point: 

\begin{equation}
\begin{array}{cl}
    \mathcal{U}(\mathcal{E}_S)_{|\mathcal{P}|=|\bar{\mathcal{P}_T}|}>\mathcal{U}(\mathcal{E}_S)_{|\mathcal{P}|=|\bar{\mathcal{P}_S}|} \\
    \frac{\partial(\mathcal{U}(\mathcal{E}_S))}{\partial|\mathcal{P}|}_{|\mathcal{P}|=|\bar{\mathcal{P}_T}|}<\frac{\partial(\mathcal{U}(\mathcal{E}_S))}{\partial|\mathcal{P}|}_{|\mathcal{P}|=|\bar{\mathcal{P}_S}|}.
\end{array}
\end{equation}

From this, we get $|\bar{\mathcal{P}_S}|>|\bar{\mathcal{P}_T}|$. 
\end{proof}

\section{C\ \ \ Extended Analysis}
\subsection{C.1 The Relationship Between GNN Size and Error}
\label{app_delta}
Here, by analyzing the relationship between GNN size and test/training error, we empirically show our tasks fall within the scope of the classical regime of generalization theory, where the test error follows the U-shaped behavior, i.e., decreases and then increases as the model size grows. From this, we can infer the satisfaction of Ineq. \tmprefinequa{inequa} in our experiments of GNNs: $\hat{\mathcal{E}}({\mathcal{D}_{n_T}^T},\bar {\mathcal{P}_S})+O\left(\sqrt{|\bar {\mathcal{P}_S}|/n_T}\right) > \hat{\mathcal{E}}({\mathcal{D}_{n_T}^T},\bar {\mathcal{P}}_T)+O\left(\sqrt{|\bar {\mathcal{P}}_T|/n_T}\right)$, where $|\bar {\mathcal{P}_S}|>|\bar {\mathcal{P}_T}|$.
This inequality suggests that smaller model sizes typically lead to lower test errors when training a model from scratch. This forms the basis for the effectiveness of our PEFT method.

\begin{figure*}[h!]
  \centering
  \subfigure[BACE]{\includegraphics[width=0.32\textwidth]{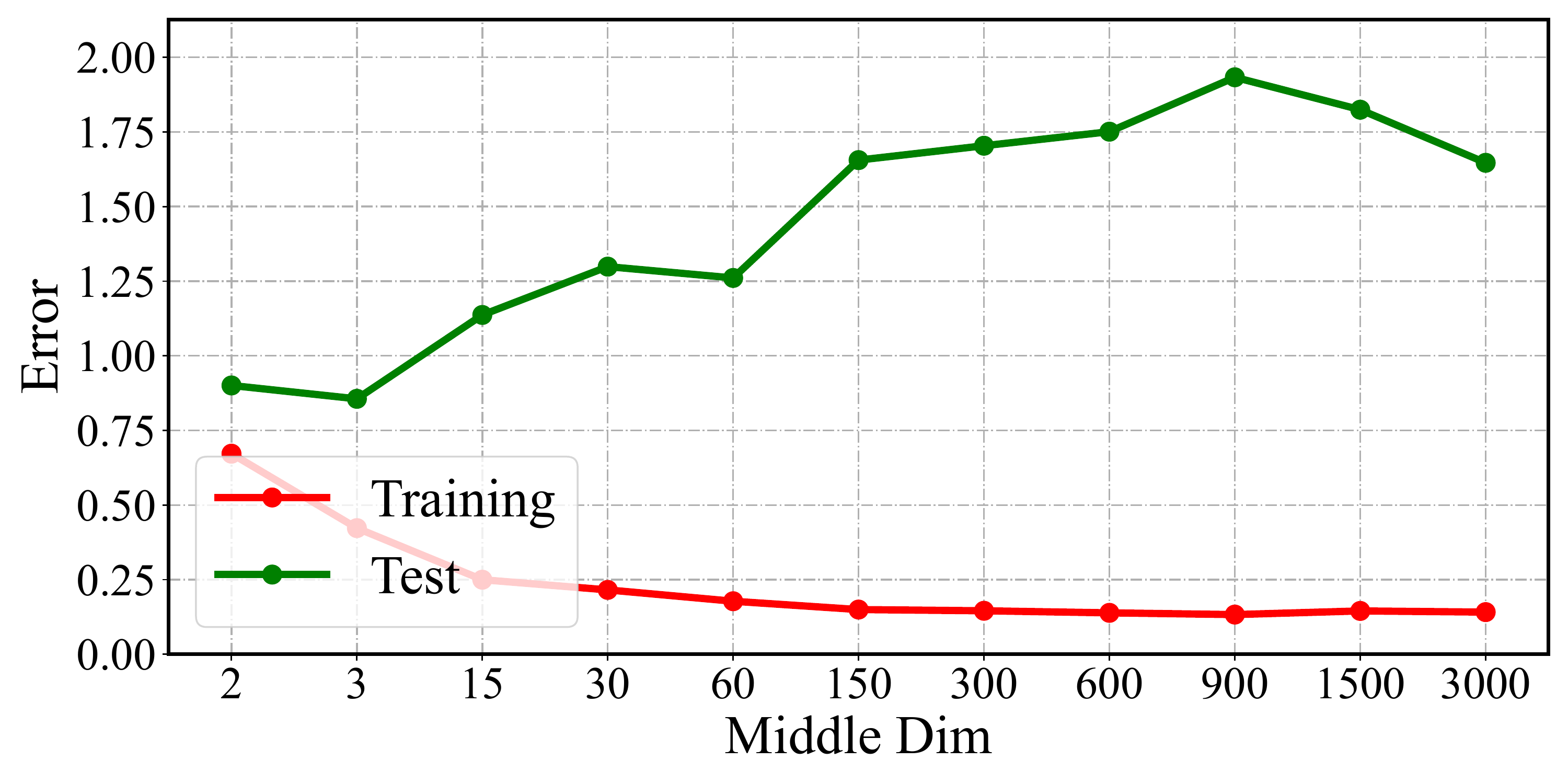}}
  \subfigure[BBBP]{\includegraphics[width=0.32\textwidth]{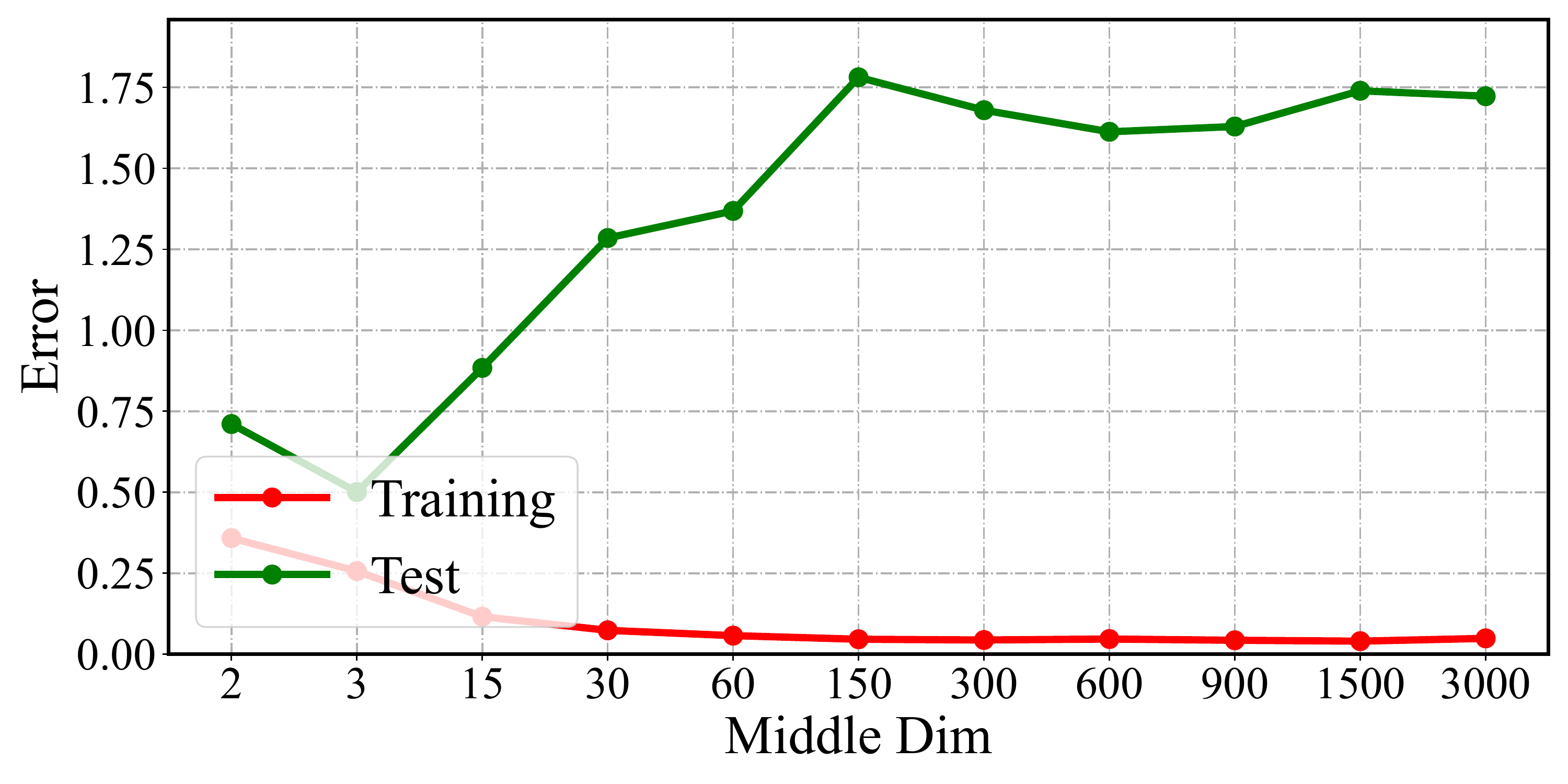}}
\subfigure[Clintox]{\includegraphics[width=0.32\textwidth]{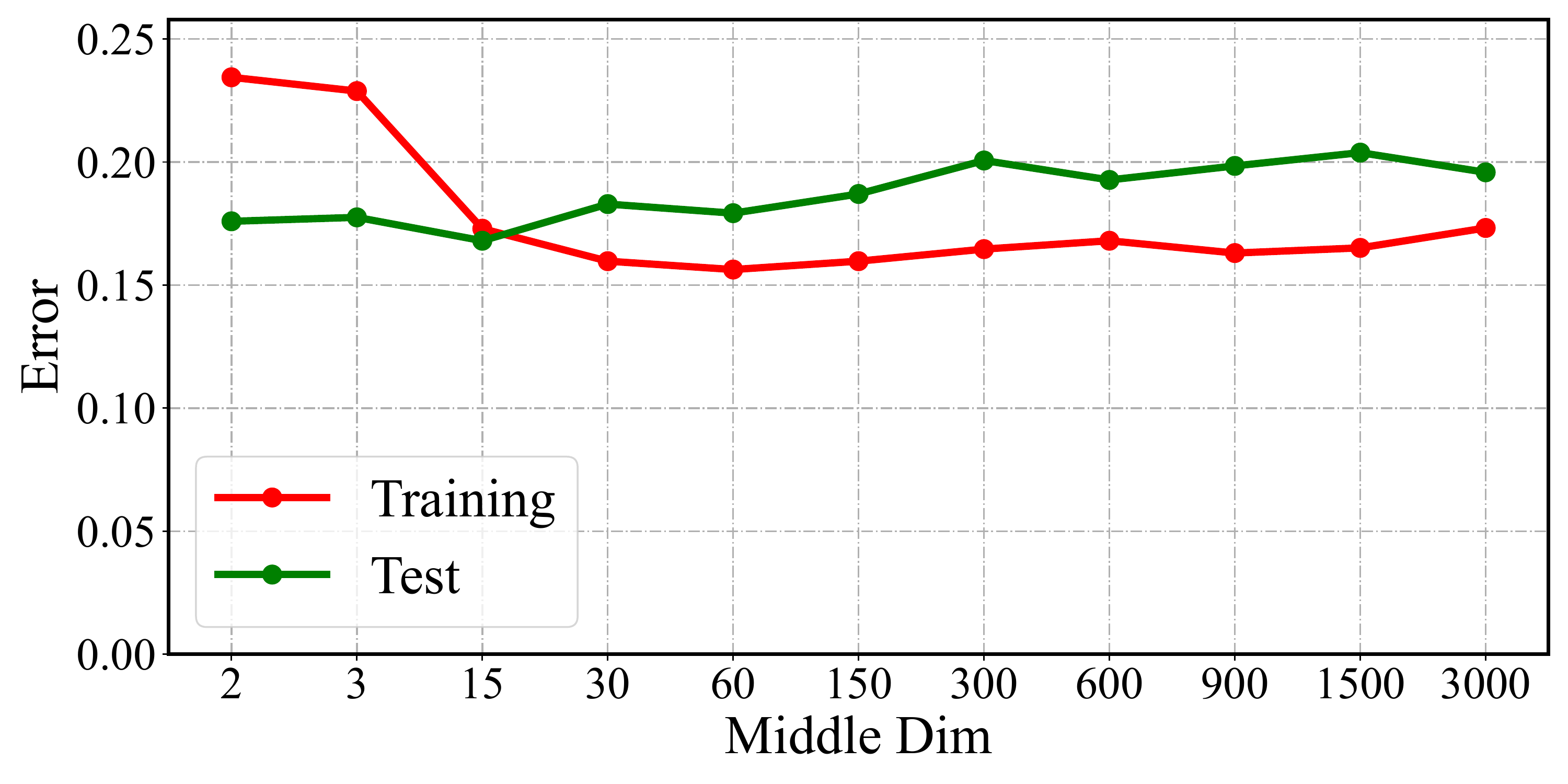}}\\
  \subfigure[SIDER]{\includegraphics[width=0.32\textwidth]{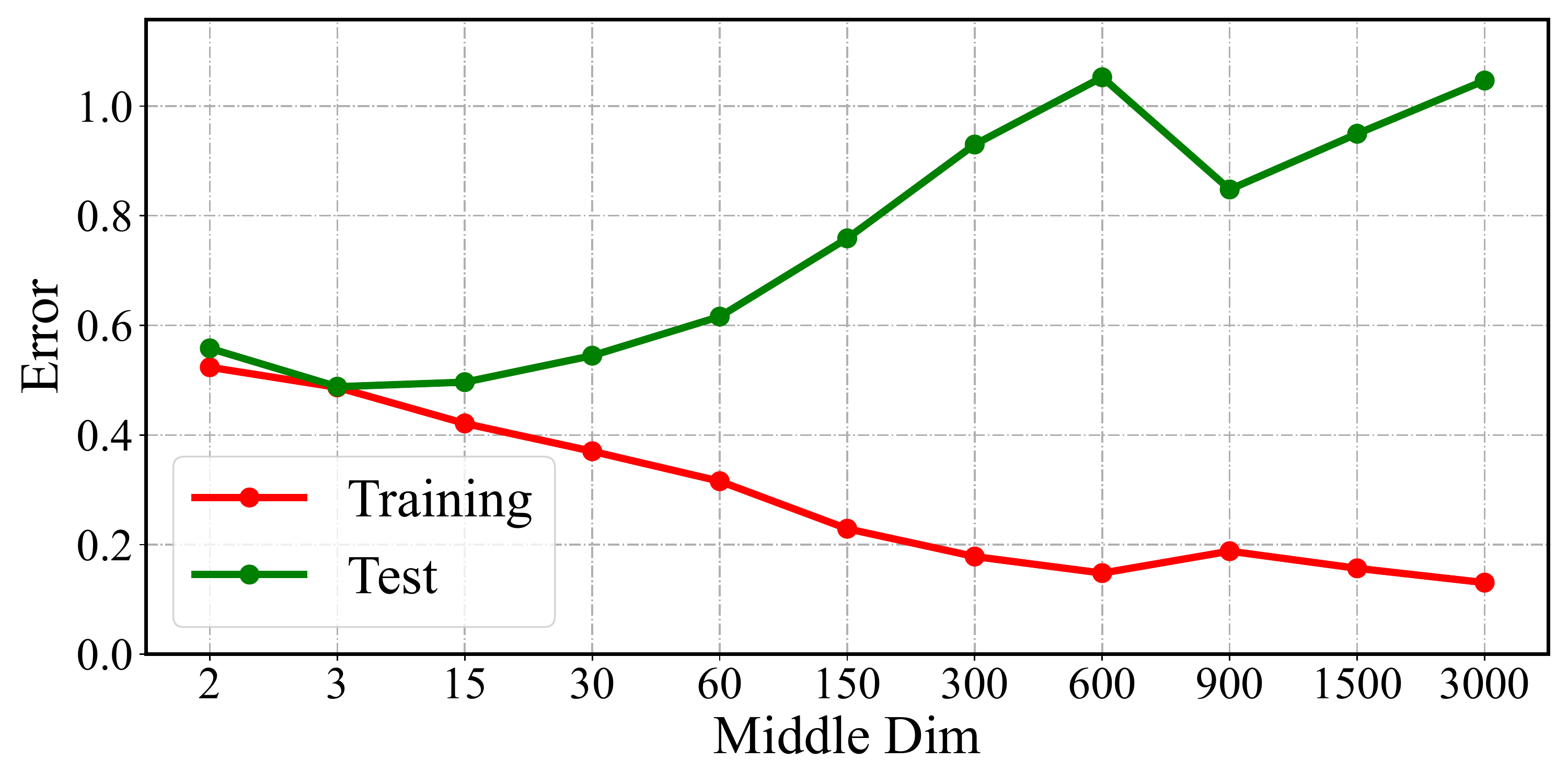}}
  \subfigure[Tox21]{\includegraphics[width=0.32\textwidth]{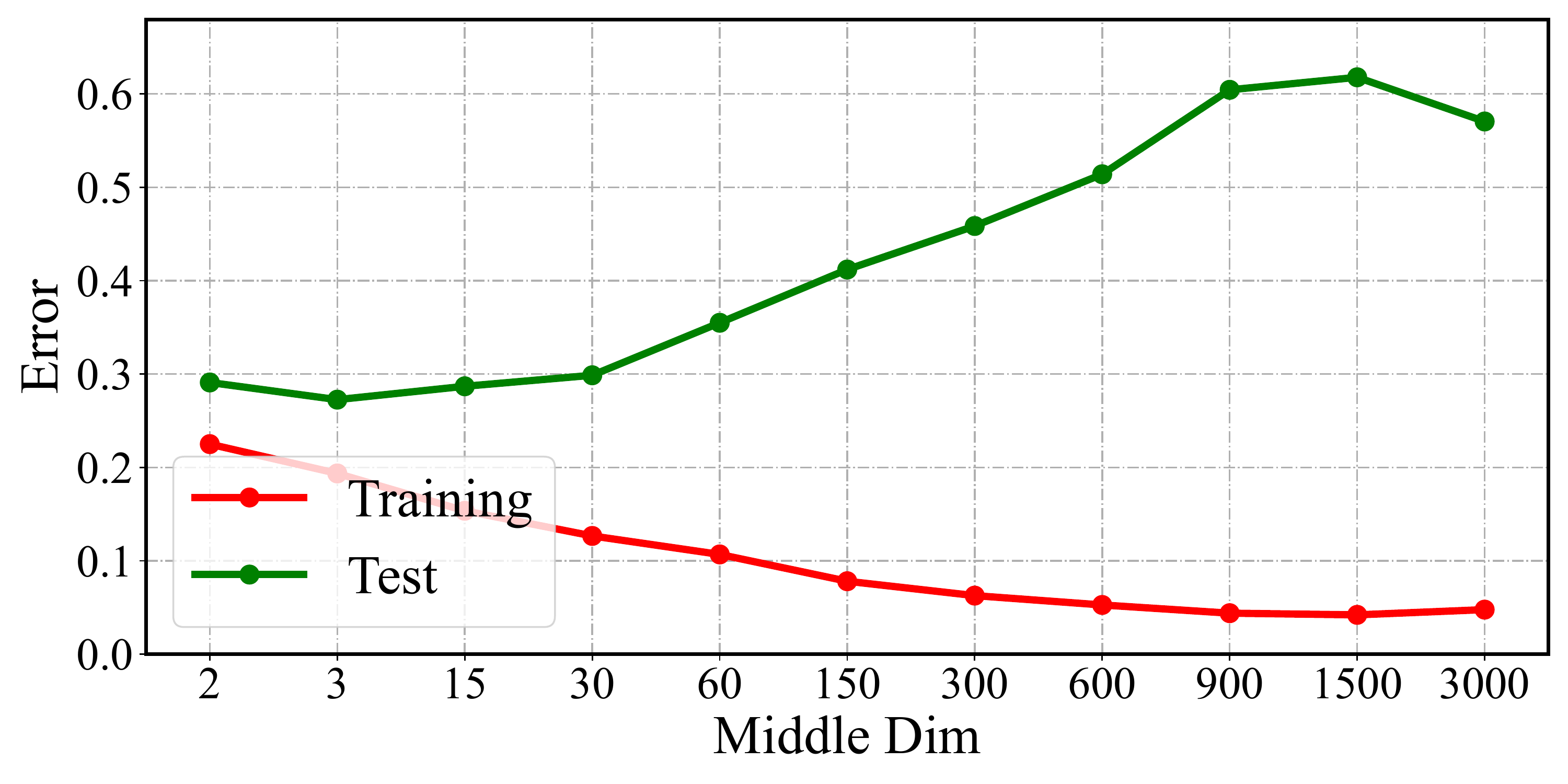}}
  \subfigure[ToxCast]{\includegraphics[width=0.32\textwidth]{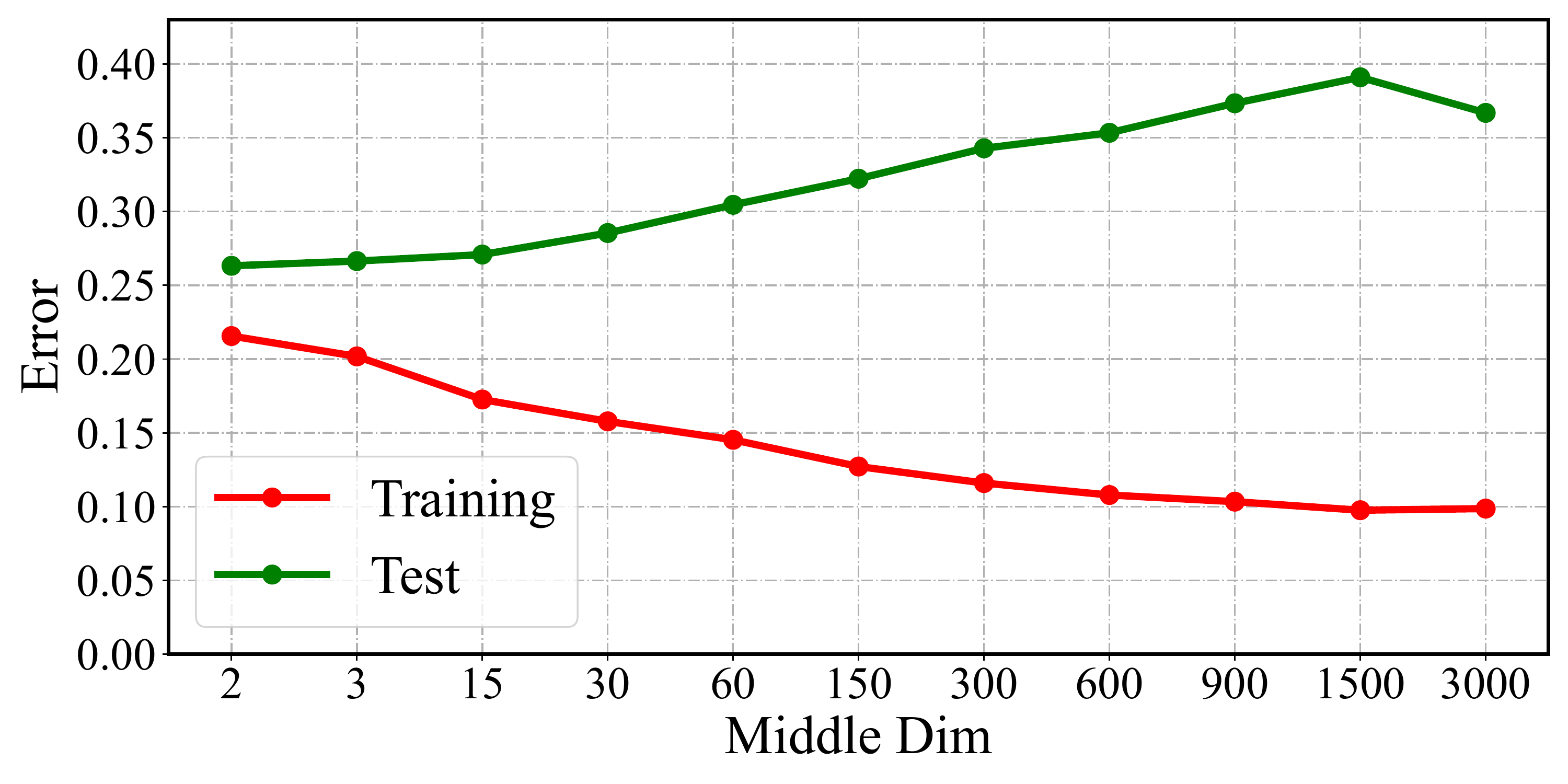}}
  \caption{Without pre-training, the training and test error across different model sizes, measured by GNN MLP middle dimensions.}
  \label{fig_error_mid}
\end{figure*}

We conducted experiments on six small molecular datasets, varying model sizes in two ways: (1) by varying embedding dimensions while fixing the MLP middle dimension at twice the embedding dimensions, and (2) by fixing embedding dimensions at 300 and varying the middle dimensions. The results are presented in Figures \ref{fig_error_emb} and \ref{fig_error_mid}. For all datasets, we made two observations. First, the test error initially decreases and then increases as the model size grows. Second, the training error does not approach zero, indicating that even a large model is not over-parameterized \cite{nakkiran2021deep}. These observations suggest that our tasks fall within the scope of the classical regime and reducing the size of the parameter space, unless too small, can help improve generalization ability.

\begin{figure*}[h!]
\addtocounter{figure}{+5}
  \centering
  \subfigure[BACE]{\includegraphics[width=0.32\textwidth]{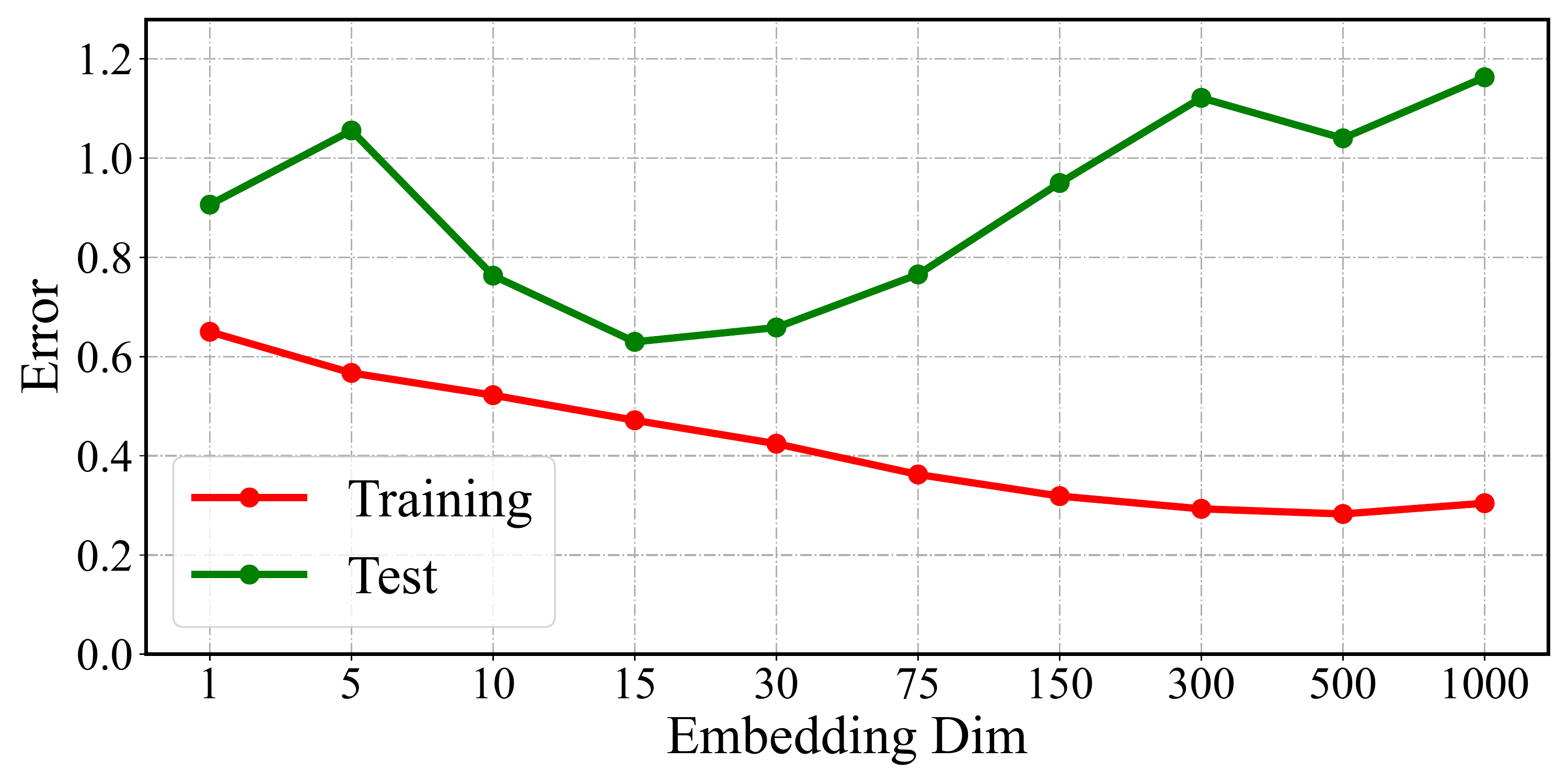}}
  \subfigure[BBBP]{\includegraphics[width=0.32\textwidth]{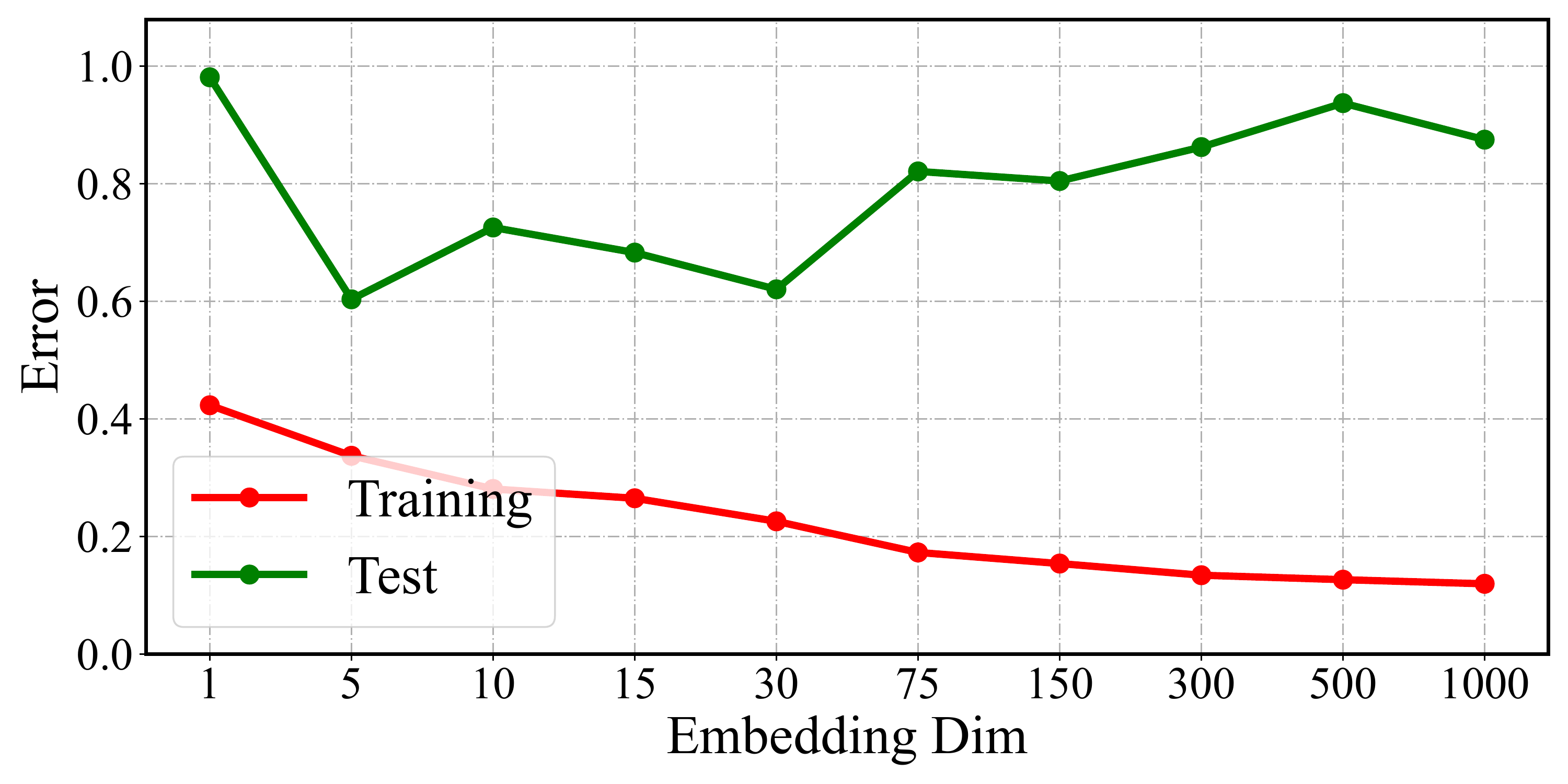}}
    \subfigure[Clintox]{\includegraphics[width=0.32\textwidth]{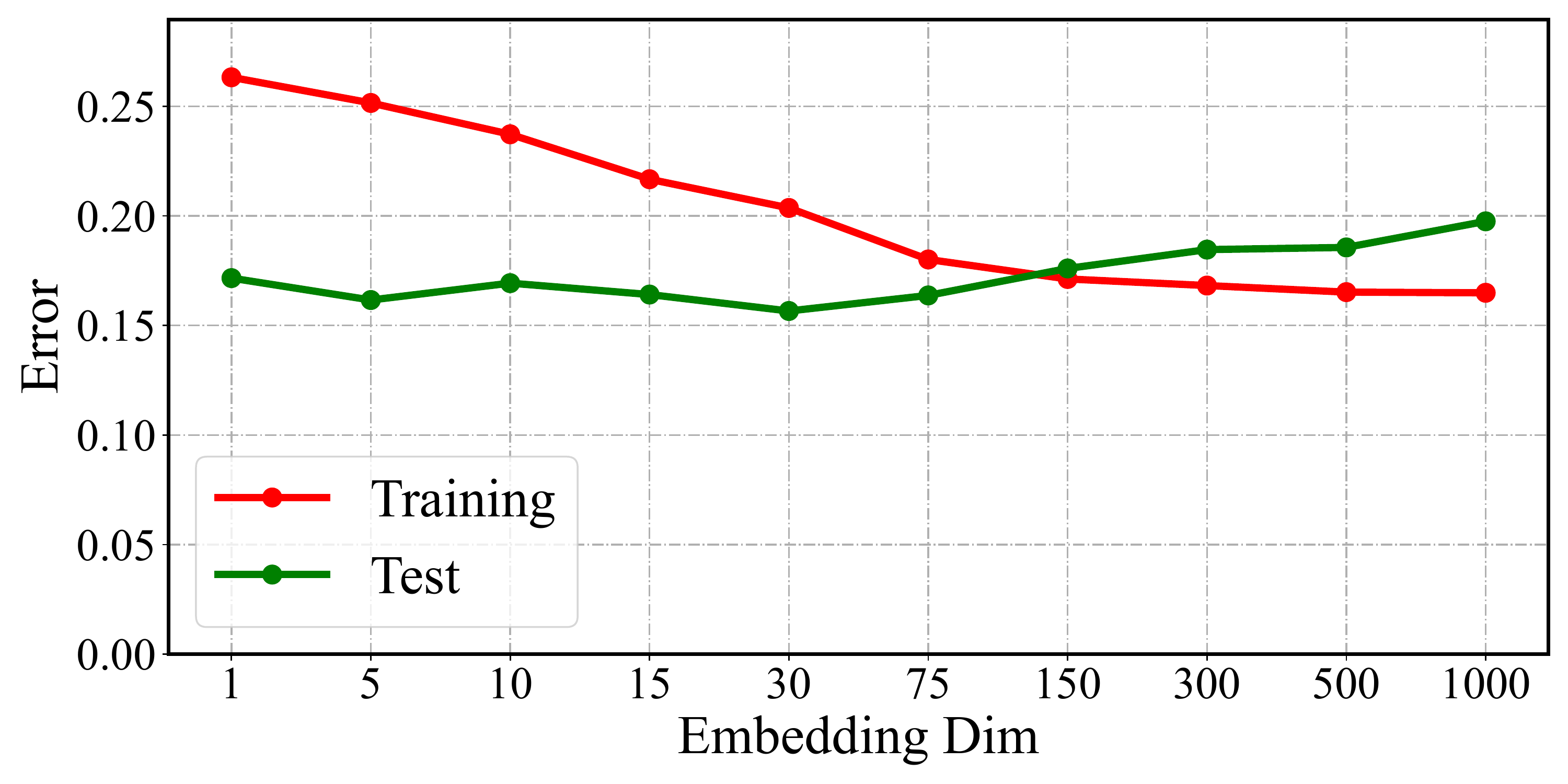}} \\
  \subfigure[SIDER]{\includegraphics[width=0.32\textwidth]{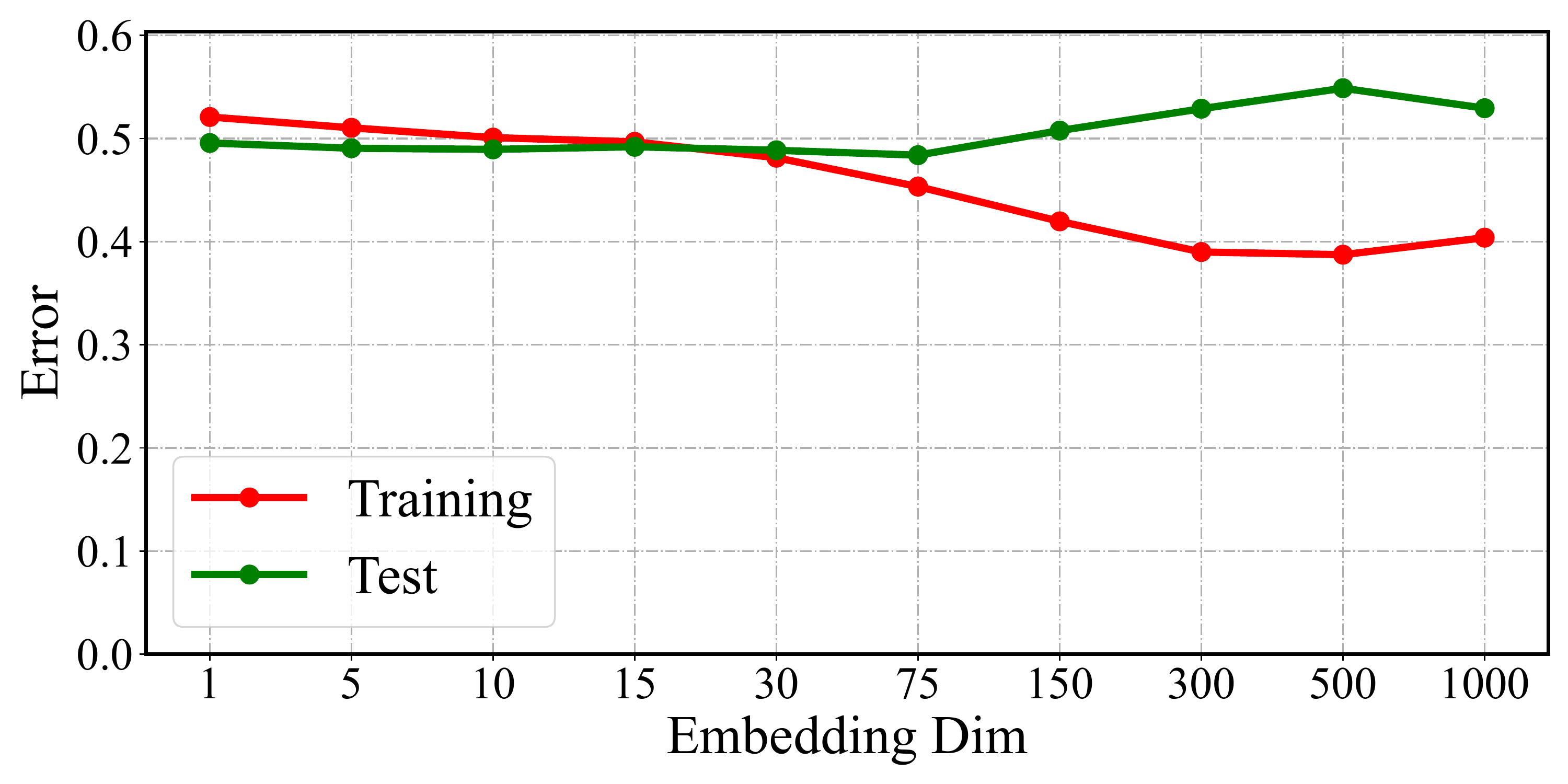}}
  \subfigure[Tox21]{\includegraphics[width=0.32\textwidth]{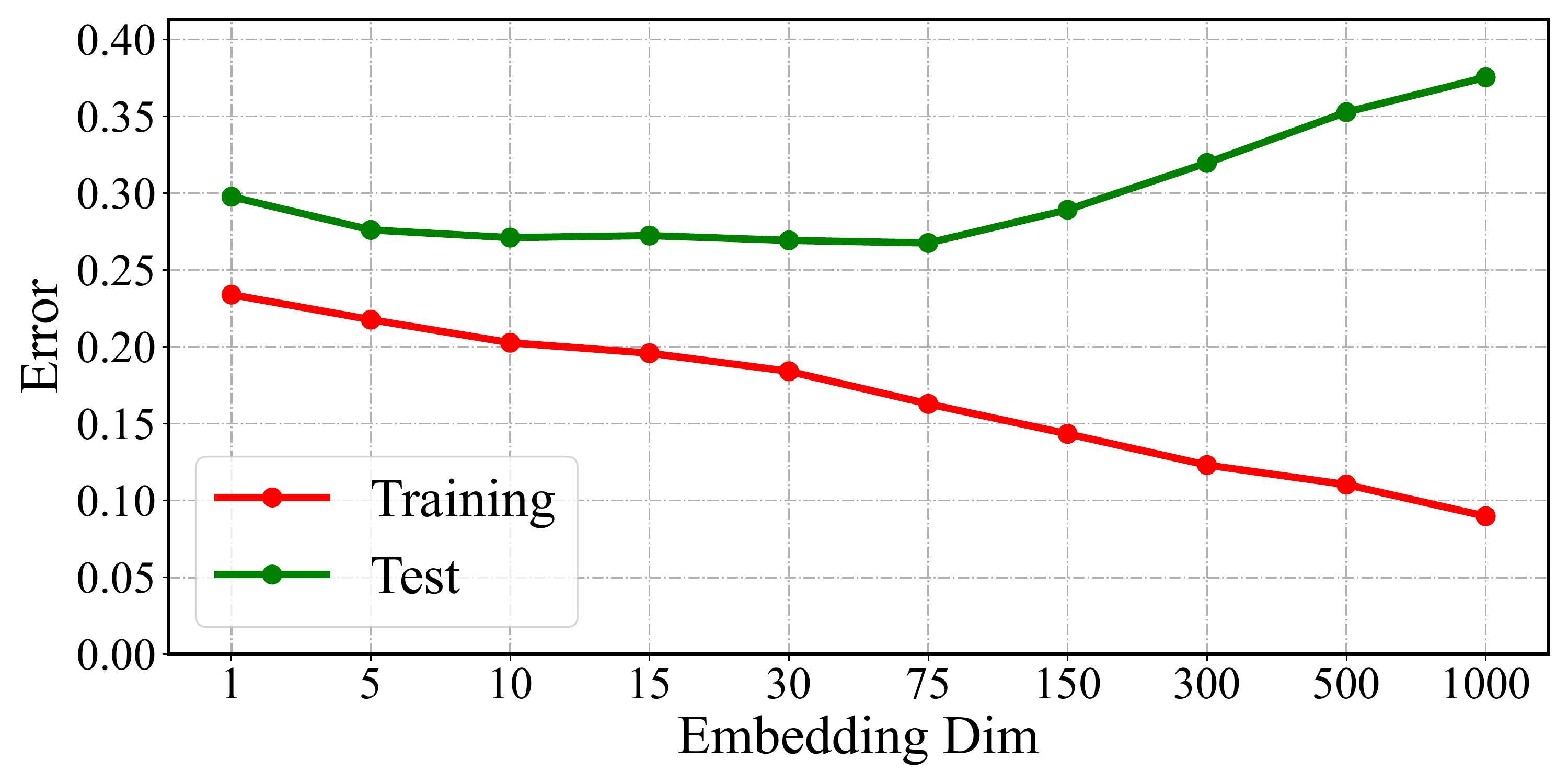}}
  \subfigure[ToxCast]{\includegraphics[width=0.32\textwidth]{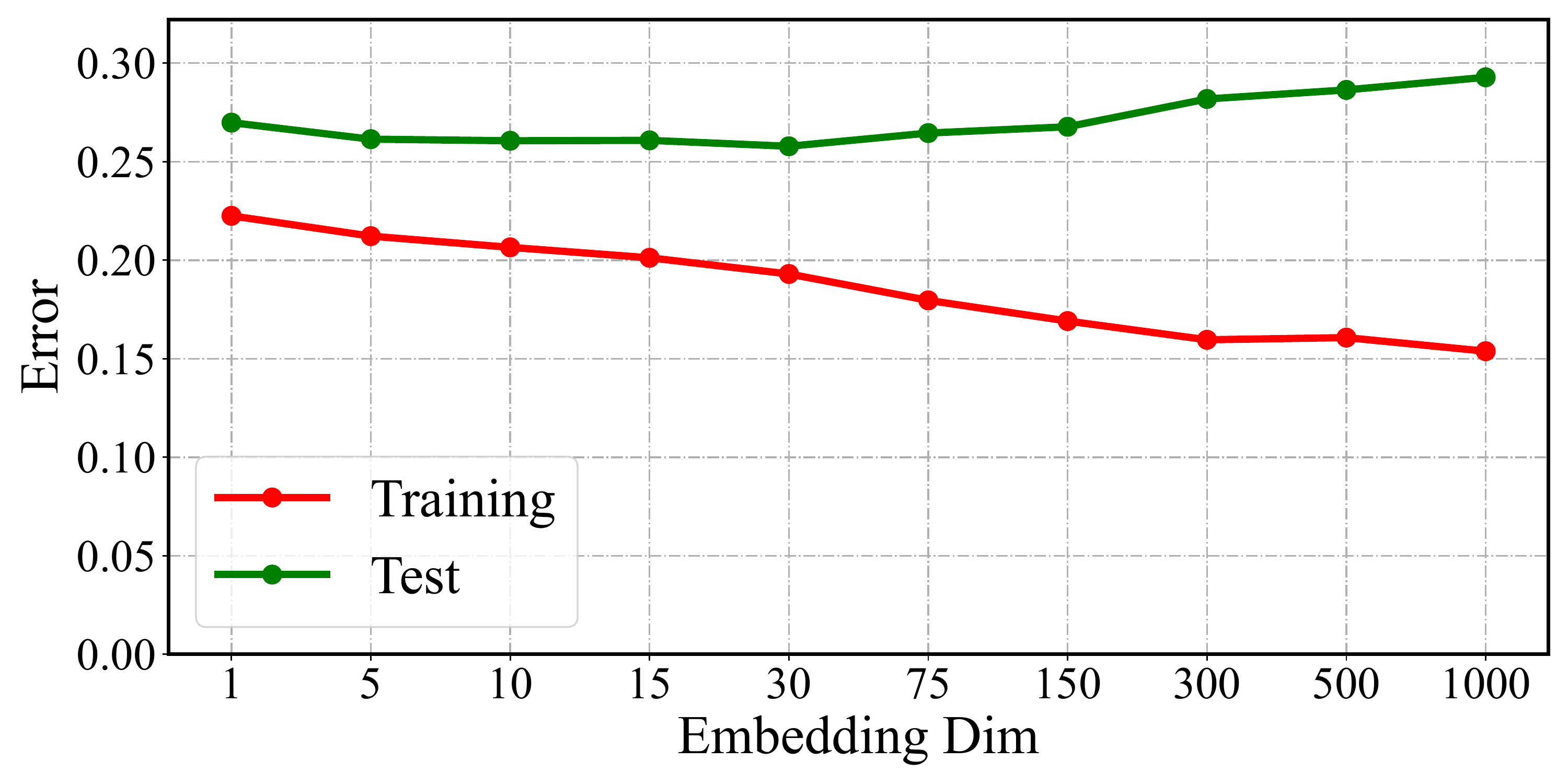}}
  \caption{Without pre-training, the training and test error across different model sizes, measured by GNN MLP embedding dimensions.}
  \label{fig_error_emb}
\end{figure*}

It is important to note that the phenomenon in question occurs exclusively in the context of training from scratch. To leverage the knowledge gained from pre-training, employing a larger model is necessary. To enhance the generalization ability of this large model, the full fine-tuning approach can be replaced with PEFT. As illustrated in Fig. \tmpreffigabl{fig_abl_1}(a), embedding dimensions between 200 to 300 exhibit the highest performance for full fine-tuning, albeit with inferior test error when training from scratch. Our AdapterGNN outperforms full fine-tuning consistently, implying that pre-training on a larger model and utilizing PEFT with fewer parameters is the optimal approach.

\subsection{C.2 How AdapterGNN Preserve Expressivity of Full Fine-tuning}
\label{app_preserve}
\begin{figure*}[ht]
  \centering
  \subfigure[BACE]{\includegraphics[width=0.32\textwidth]{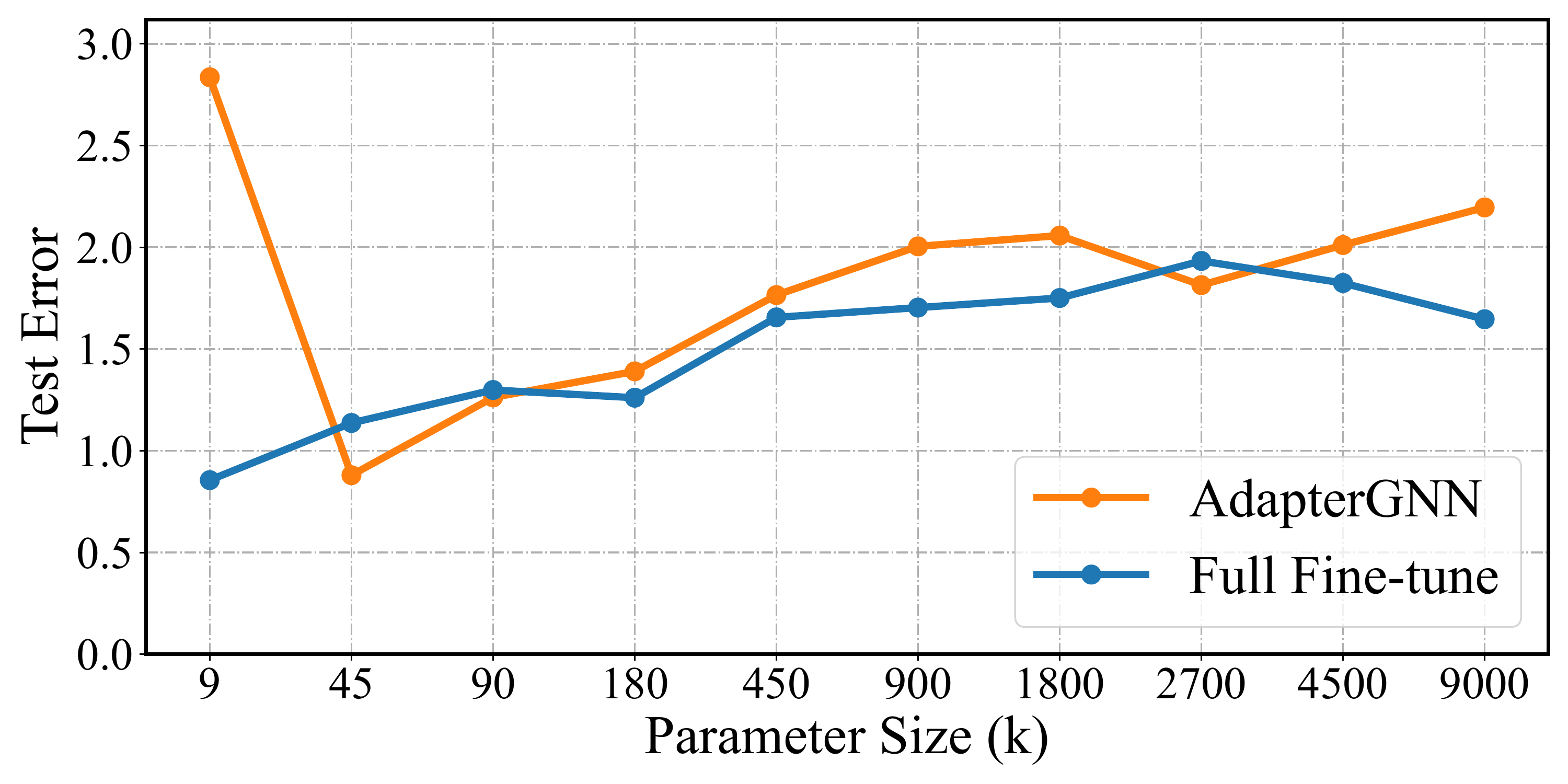}}
  \subfigure[BBBP]{\includegraphics[width=0.32\textwidth]{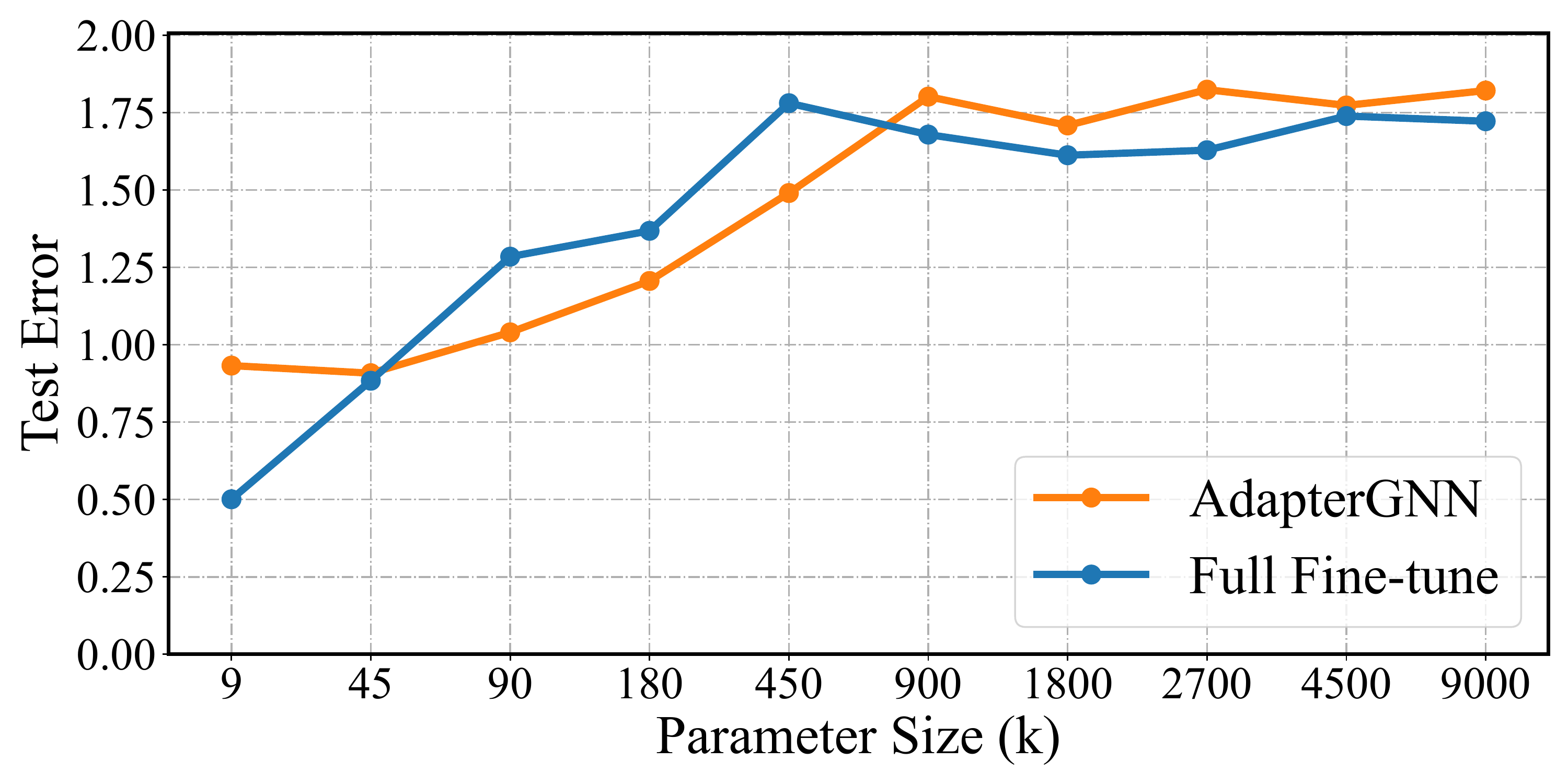}}
    \subfigure[Clintox]{\includegraphics[width=0.32\textwidth]{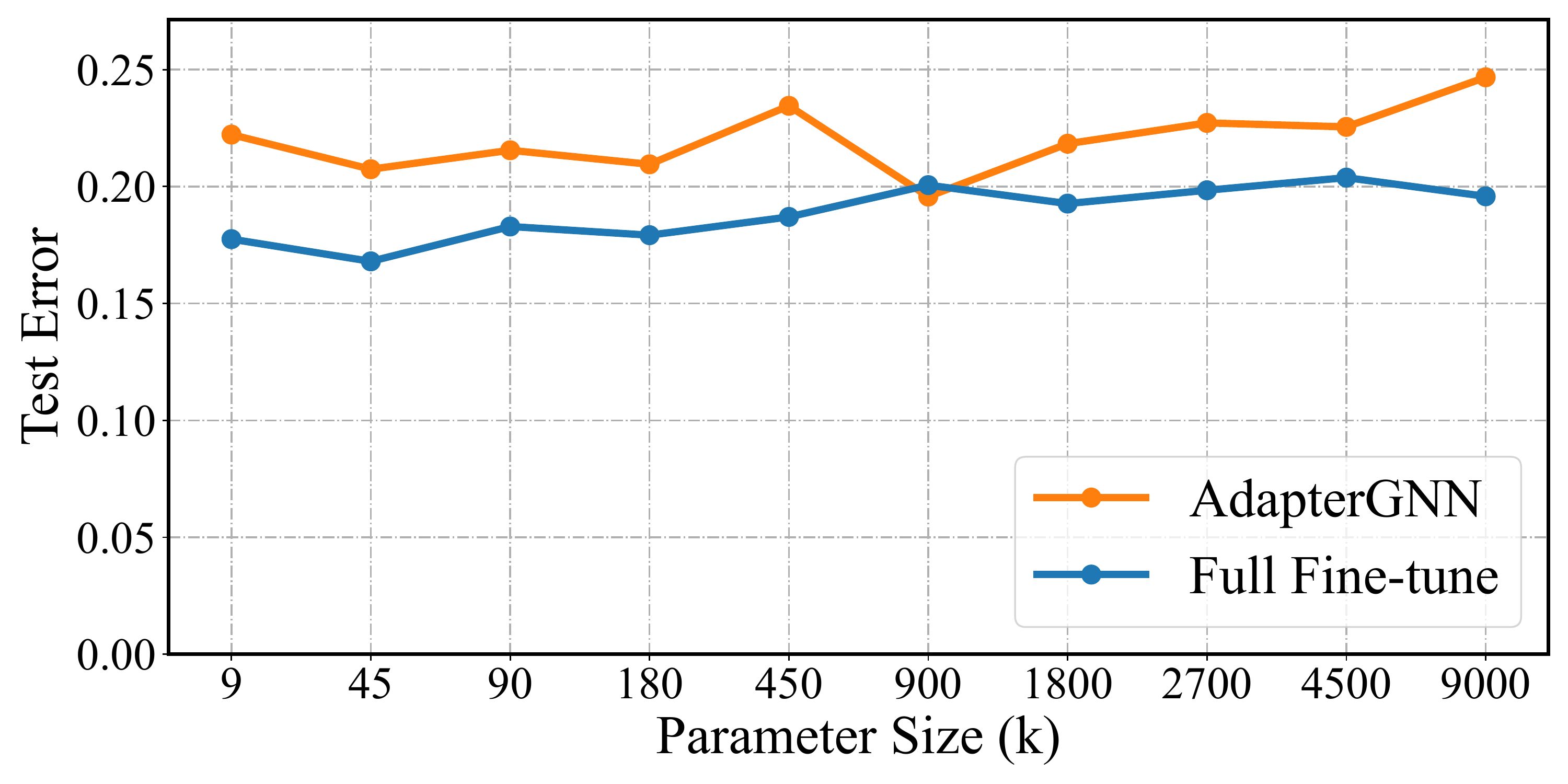}}
    \\
  \subfigure[SIDER]{\includegraphics[width=0.32\textwidth]{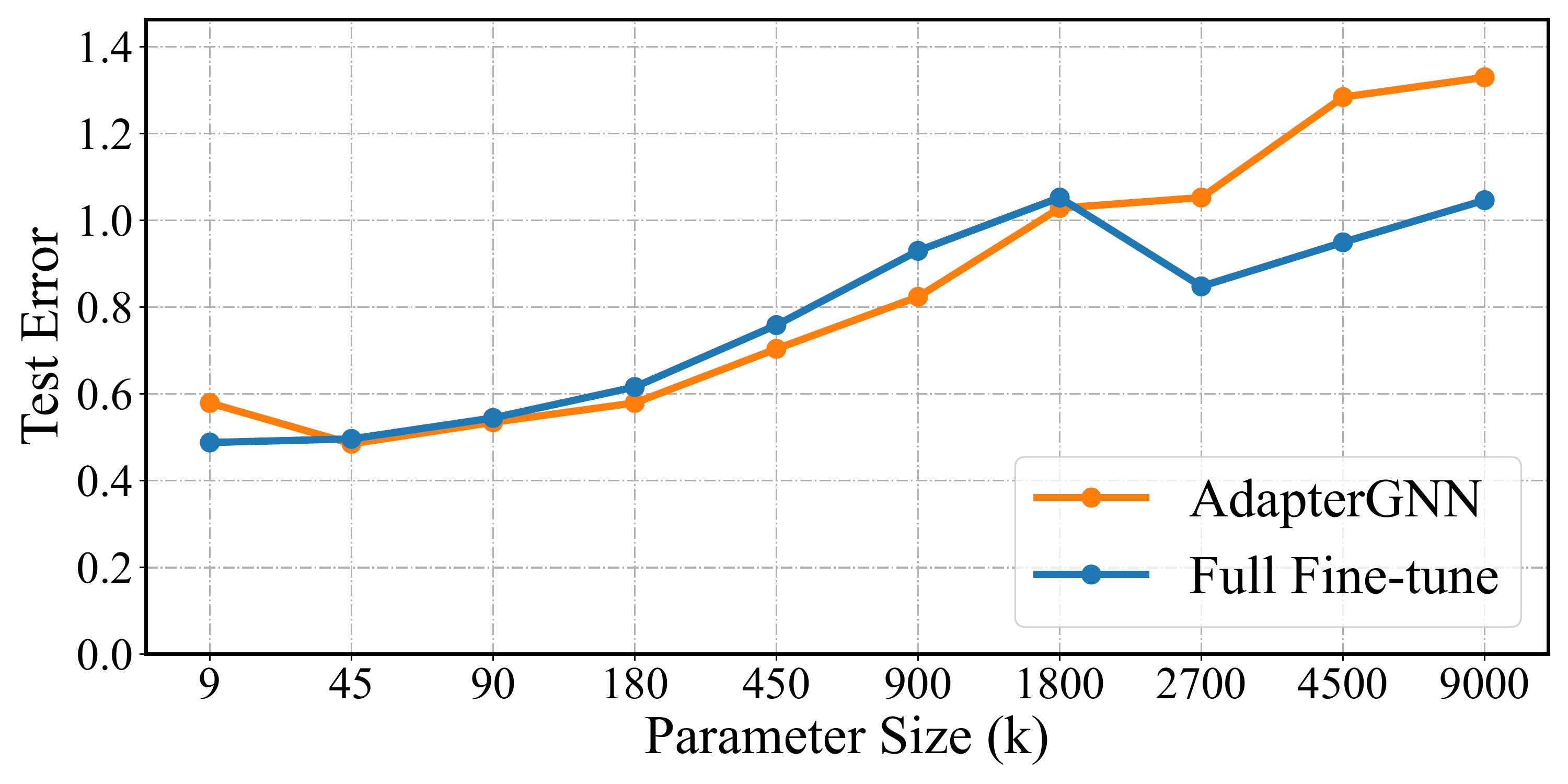}}
  \subfigure[Tox21]{\includegraphics[width=0.32\textwidth]{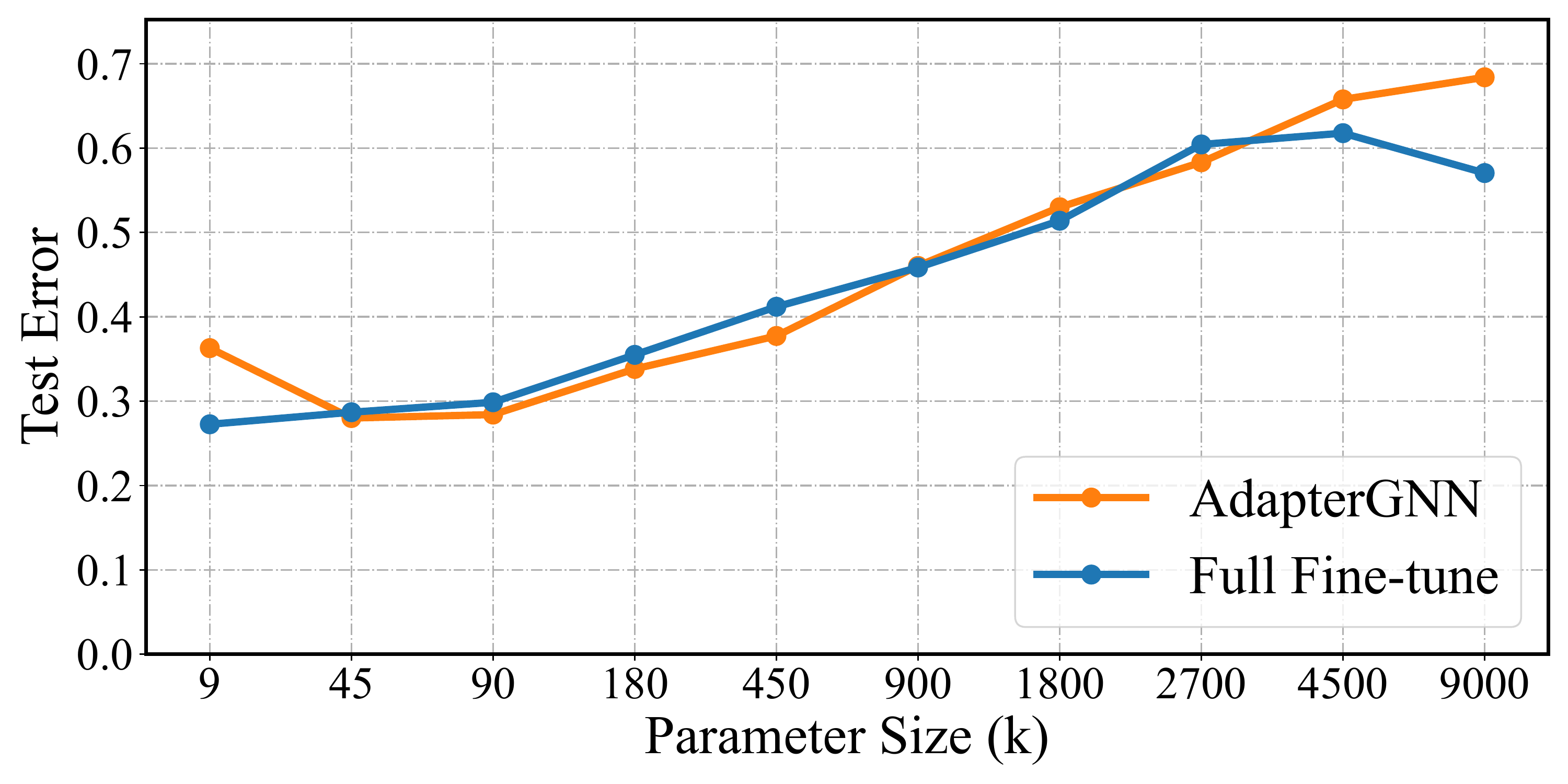}}
  \subfigure[ToxCast]{\includegraphics[width=0.32\textwidth]{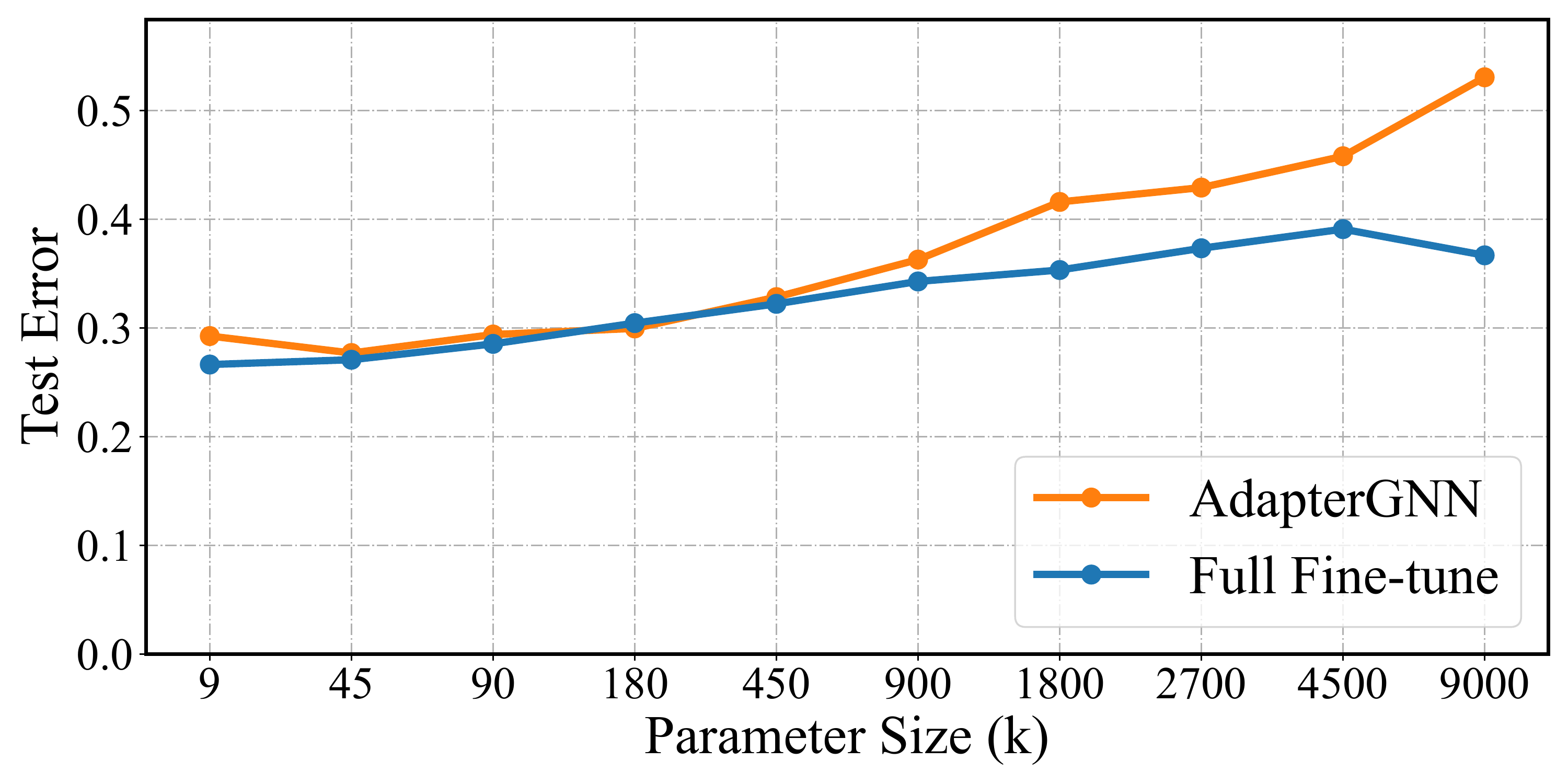}}
  \caption{Without pre-training, comparison of test errors for AdapterGNN and full fine-tuning across different tunable parameter sizes.}
  \label{fig_expressivity}
\end{figure*}

Our specially designed AdapterGNN maintains the expressivity of full fine-tuning, which is a crucial prerequisite for its superior generalization ability. To verify the comparable expressivity under varying structures, we perform empirical experiments by training both the structure of AdapterGNN (with GNN MLPs of AdapterGNN remaining fixed) and the original GNN from scratch. We modify the bottleneck dimension of AdapterGNN and the middle dimension of the original GNN MLP to alter their parameter sizes. Subsequently, we compare their test errors under different tunable parameter sizes, with the results presented in Figure \ref{fig_expressivity}. These findings indicate that given equivalent tunable parameter sizes, the test error of AdapterGNN is nearly equivalent to that of full fine-tuning in most cases, thereby confirming that the structure of AdapterGNN successfully preserves the expressivity of the original GNN structure.

\subsection{C.3 Analysis on the Training and Inference FLOPs and Latency}
\label{app_flops}
\begin{figure}[ht]
    \centering
    \includegraphics[width=0.45\textwidth]{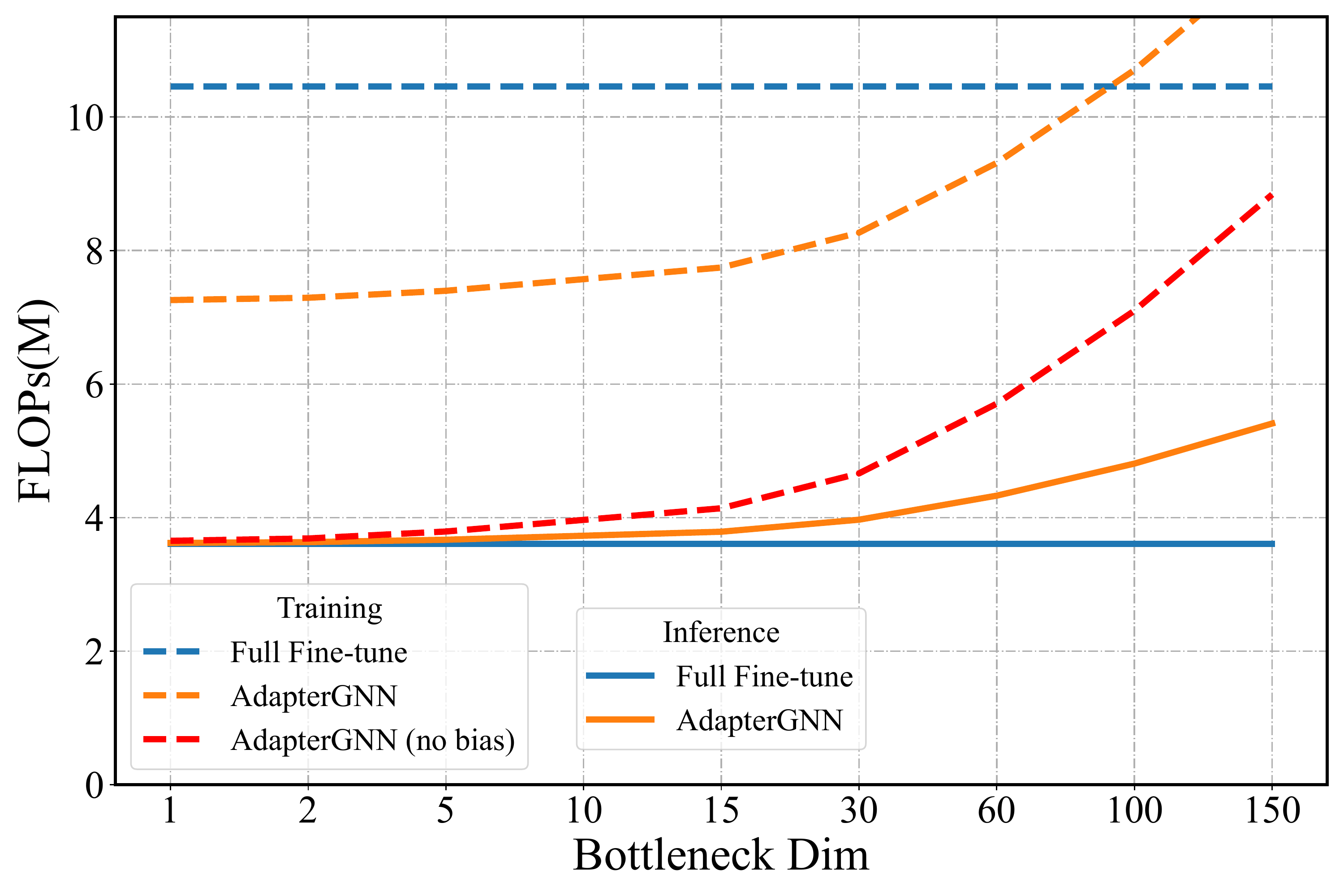}
  \caption{Comparison of training and inference FLOPs.}
  \label{fig_flops}
\end{figure}

In addition to parameter efficiency, we also analyze the computational efficiency. We first measure the FLOPs (floating point operations) during a single training/inference batch iteration. We conducted this analysis by varying the bottleneck dimension of AdapterGNN and comparing two versions: the first version has a tunable GNN MLP bias, which is practically adopted to improve performance, and the second version does not have a tunable bias, making it more efficient. 
Figure \ref{fig_flops} presents the results of this analysis.
For training, compared with full fine-tuning, AdapterGNN requires fewer back-propagation gradient computations, resulting in fewer FLOPs, especially when the bottleneck dimension is small. Moreover, the version without bias requires even fewer FLOPs.  
During inference, both versions of AdapterGNN require the same FLOPs, and the extra modules inserted in AdapterGNN result in slightly more FLOPs than full fine-tuning. 

We also measure latency during a single training/inference batch iteration using NVIDIA A100 GPUs with Pytorch \textit{torch.jit.script} multi-process mechanism, which maximizes the parallel computing ability of AdapterGNN. The bottleneck dimension of 15 is chosen as in our main experiments. The results in Table \ref{tab_latency} demonstrate that both versions of AdapterGNN can improve training speed, with the version with bias improving it by 11.5\%, and the version without bias by 12.8\%. We note that AdapterGNN only brings slight improvements in training speed because Pytorch can parallelly compute FLOPs within the same tensor, making full fine-tuning fast enough.
For inference, additional FLOPs cost more latency, with the version with bias adding 11.1\% and the version without bias adding 7.2\%.

\begin{table}
\addtocounter{table}{+5}
    \centering
    \caption{Comparison of training and inference latency.}
    \begin{tabular}{ccc}
    \toprule
      \multirow{2}{*}{Method} & \multicolumn{2}{c}{Latency(ms)}\\
      ~ & Training & Inference \\
    \midrule
    Full Fine-tuning &14.8 & 2.50  \\
    AdapterGNN & 13.1 & 2.77 \\
    AdapterGNN (no bias)  & 12.9 & 2.68 \\
    \bottomrule
    \end{tabular}
    \label{tab_latency}
\end{table}

\section{D\ \ \ Experiment Details}
\subsection{D.1 Implementations of AdapterGNN}
\label{app_imple_detail}
PyTorch and PyG \cite{Fey/Lenssen/2019} are used to conduct all experiments on NVIDIA A100 GPUs in this work. All hyperparameters and training strategies are consistent with the previous work \cite{hu2019strategies}. The data is split into the out-of-distribution scaffold and species splits, which are also consistent. For full fine-tuning, the default embedding dimension is set to 300. For the AdapterGNN hyperparameters, the bottleneck dimension is set to 15, and the starting value of the learnable scaling is set to 0.01. Our results are achieved \textbf{WITHOUT} any special hyperparameter tuning for individual datasets to ensure efficiency. In practice, we also tune the bias of the original MLPs as BitFit \cite{zaken2021bitfit}. It keeps efficiency and can provide additional improvement.

\begin{table*}[h!]
  \caption{Statistics of datasets for downstream tasks.}
  \centering
  \begin{tabular}{cccccccccc}
    \toprule
      \textbf{Dataset} & BACE &	BBBP	&ClinTox	&HIV	&Sider&	Tox21&	MUV	&ToxCast &PPI\\
      \midrule
    \textbf{\# Graphs} & 1513&	2039&	1478&	41127&	1427&	7831& 93087&	8575& 88k \\
    \textbf{\# Tasks} & 1&	1&	2&	1&	27&	12& 17&	617& 40 \\
    \bottomrule
  \end{tabular}
  \label{table_dataset}
\end{table*}

\subsection{D.2 Implementation Details of Other PEFT Techniques in GNNs}
\label{app_implementation}

Recently, researchers have proposed several PEFT methods. However, most of them are only applied to the transformer-based model in the NLP field. Here, we will detail how we utilize these methods in GNNs.

\paragraph{BitFit \cite{zaken2021bitfit}.} We tune all bias parameters of each linear layer in each GNN MLP. This technique brings slight improvement in performance.

\paragraph{(IA)$^3$ \cite{liu2022few}.} In each GNN MLP, we modify the input of the linear layer by incorporating learnable weights. Specifically, if the embedding dimension is denoted as $n_{ebd}$, then the input to the linear layer is of shape $n_{batch} \times n_{ebd}$. We design learnable weights of shape $n_{ebd}$ and reweight the input by element-wise multiplication with these weights before feeding it into the linear layer. This approach is both efficient and effective. However, due to the limited number of tunable parameters (only 0.24\% of the total), the expressivity of the model is still restricted.

\paragraph{LoRA \cite{hu2021lora}.} LoRA modifies the output of the linear layer in each MLP. To achieve this, it uses a parallel module that consists of two sequential linear layers with a bottleneck dimension. The module takes the same input as the original linear layer and adds its output to the original output. By using a small bottleneck, the size of tunable parameters is reduced from $n_{in}\times n_{out}$ to $n_{in} \times n_{bottleneck} + n_{bottleneck} \times n_{out}$. Furthermore, after PEFT, the tuned linear layers can be multiplied and directly added to the original frozen linear layer. This process results in no extra parameters during inference. Although LoRA achieves higher performance with more parameters, it is still no better than full fine-tuning.

\paragraph{Prompt.} Prompt tuning is a widely acknowledged technique in the NLP field, which involves modifying the input of the network while keeping the parameters fixed. In GNNs, there are two types of prompt tuning techniques: feature prompt (feat) and node prompt (node). With the feature prompt, a learnable feature is added to the node embedding. In addition to previous work \cite{fang2022prompt}, we also tune batch normalization. For the node prompt, a fully connected learnable virtual node is added to each GNN layer. Prompt tuning is a highly efficient technique, but its performance in GNNs is not as good as in NLP. We conjecture that this is due to the gap between GNNs and transformer-based models. In GNNs, modifying the input alone may not achieve the same level of expressivity as in NLP \cite{he2021towards}, which could be one of the reasons for the performance gap.

\paragraph{Adapter \cite{houlsby2019parameter}.} Similar to LoRA, the adapter also utilizes two sequential linear layers with a bottleneck to add its output to the original output. Additionally, an activation layer is inserted to form an adapter, and an additional batch normalization layer is added behind it. We evaluate two types of adapters: sequential (seq) and parallel (par). The sequential adapter takes the output of each GNN layer as input and adds its output to the original GNN layer output in the end. The parallel adapter takes the input before GNN MLP (or before message passing) as input and adds its output to the original output of GNN MLP. Notably, this is also different from LoRA. The adapter is parallel to GNN MLP, while LoRA is parallel to the linear layer of GNN MLP. With these advanced techniques, the adapter has improved expressivity and achieves superior performance, even outperforming full fine-tuning with only 5\% of the total parameters. Therefore, our AdapterGNN primarily adopts several techniques from adapters.

\paragraph{Partial.} We have implemented a partial fine-tuning approach for our GNN layers. Specifically, Partial-1 indicates that only the last layer is fine-tuned. This technique is not quite efficient.

\subsection{D.3 Details of Datasets}
\label{app_dataset}

For pre-training, the dataset of the chemistry domain contains 2 million unlabeled molecules sampled from the ZINC15 \cite{sterling2015zinc} database. The dataset of the biology domain contains 395K unlabeled protein ego-networks from PPI networks.

For downstream fine-tuning, we utilize eight binary classification datasets published for downstream tasks in the chemistry domain. For the biology domain, we use 88K labeled protein ego-networks from PPI networks to form 40 binary classification tasks to predict fine-grained biological functions. These datasets are adopted in Hu.'s work \cite{hu2019strategies}. We provide detailed statistics of our downstream datasets in Table \ref{table_dataset}. The first eight datasets pertain to the chemistry domain, while the last PPI dataset is related to the biology domain.

\subsection{D.4 Details of Pre-trained GNN Models}
\label{app_pretrain}
To ensure a fair comparison, we conducted experiments following the same settings outlined in the previous work \cite{hu2019strategies}. We utilized a 5-layer GIN backbone, open-source pre-trained checkpoints, and the same fine-tuning hyperparameters. We adopt five different pre-training strategies. Specifically, EdgePred \cite{hamilton2017inductive} masks and reconstructs edges to predict edge existence. 
Hu.’s work \cite{hu2019strategies} proposes AttrMasking and ContextPred which can explore graph attributes and structure. GraphGL \cite{you2020graph} proposes a graph contrastive learning framework for learning unsupervised representations of graph data. SimGRACE \cite{xia2022simgrace} takes the GNN model with its perturbed version as two encoders to obtain two correlated views for contrastive learning without data augmentations. For the latter two, we used the default checkpoints available in their published repositories without grid searching for fair and efficient comparison.

\end{document}